\documentclass[10pt]{article}

\usepackage{subfigure}
\usepackage{booktabs,multirow}
\usepackage{amsmath,amsfonts,amssymb}
\usepackage{graphicx,epsfig,epsf,xspace,setspace}
\usepackage{verbatim}
\usepackage{url}
\usepackage[hidelinks]{hyperref}
\usepackage{latexsym}
\usepackage{multicol, multirow, color}
\usepackage[dvipsnames]{xcolor}
\usepackage{natbib}
\setcitestyle{numbers,sort&compress,square}
\usepackage[margin=1in]{geometry}
\usepackage{etoolbox}

\usepackage{macros}

\begin{document}

\title{Language Models in the Loop: \\
Incorporating Prompting into Weak Supervision}
\date{}
\author{Ryan Smith\thanks{Equal Contribution} \thanks{Snorkel AI} \and Jason A. Fries\footnotemark[1] \footnotemark[2] \thanks{Stanford University} \and Braden Hancock\footnotemark[2] \and Stephen H. Bach\footnotemark[2] \thanks{Brown University} \and
\\
\small {\tt \{ryan.smith, jason.fries, braden, steve\}@snorkel.ai}}

\maketitle

\AtBeginEnvironment{tabular}{\footnotesize}

\begin{abstract}
We propose a new strategy for applying large pre-trained language models to novel tasks when labeled training data is limited.
Rather than apply the model in a typical zero-shot or few-shot fashion, we treat the model as the basis for labeling functions in a weak supervision framework.
To create a classifier, we first prompt the model to answer multiple distinct queries about an example and define how the possible responses should be mapped to votes for labels and abstentions.
We then denoise these noisy label sources using the Snorkel system and train an end classifier with the resulting training data.
Our experimental evaluation shows that prompting large language models within a weak supervision framework can provide significant gains in accuracy.
On the WRENCH weak supervision benchmark, this approach can significantly improve over zero-shot performance, an average 19.5\% reduction in errors.
We also find that this approach produces classifiers with comparable or superior accuracy to those trained from hand-engineered rules.
\end{abstract}

\section{Introduction}

Large pre-trained language models~\citep{devlin:naacl19, brown:neurips20, raffel:jmlr20, gao:arxiv20, rae:arxiv21} have shown remarkable zero-shot and few-shot performance on a range of natural language tasks.
By prompting them to answer queries, users can tap vast knowledge acquired through large-scale self-supervised pre-training.
Prompting~\citep{liu:arxiv21} refers to the emerging practice of conditioning a language model on an input representing a query and interpreting the output as a solution to the task.
For example, in a web spam classification task, we could give the prompt ``The following comment is spam. Yes or No? Subscribe to my channel! example.com/12345'' and compute whether the continuation ``Yes'' or ``No'' is more probable to make a prediction.
Remarkably, large pre-trained models can generalize in non-trivial ways to unseen tasks~\citep{brown:neurips20, mishra:acl22,sanh:iclr22, wei:iclr22}.
Beyond being useful for solving tasks directly, pre-trained language models are instances of foundation models~\citep{bommasani:arxiv21}, large pre-trained models that can be used as the foundation for new models that are better suited to specialized tasks, either because they are more accurate, less computationally expensive, or both.
Building on top of foundation models is an important challenge for data science, as data scientists often need to create predictive models, particularly from limited labeled training data.
In this work, we investigate how to direct the knowledge contained in pre-trained language models toward the creation of labeled training data for models that generalize beyond the performance of the source language model.

\begin{figure}[t]
    \centering
    \includegraphics[width=1.0\textwidth]{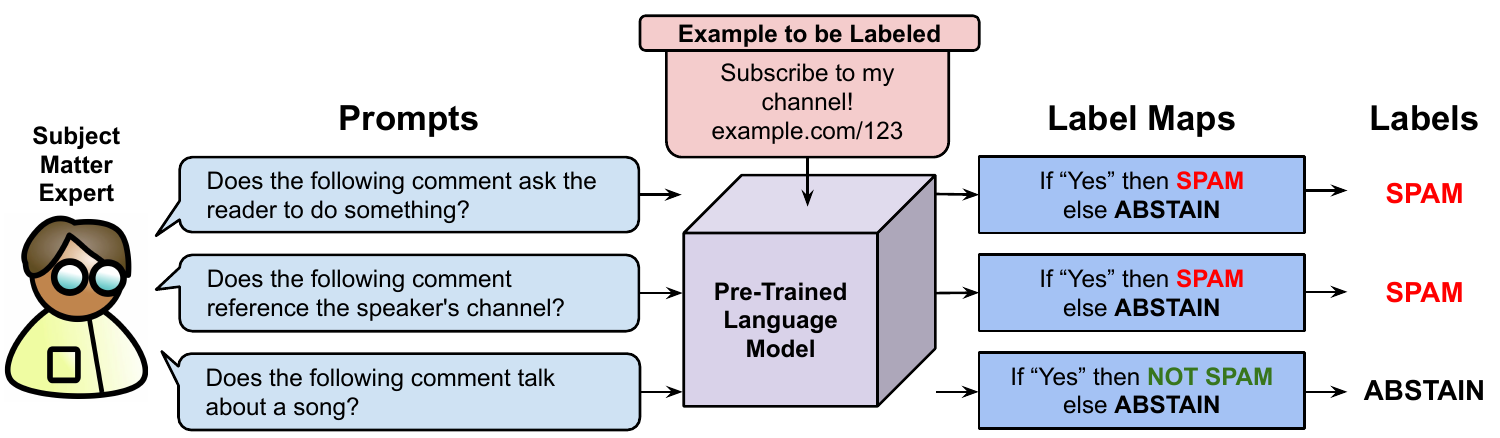}
    \caption{An overview of how a subject matter expert (SME) can use prompting to create weak supervision sources. The SME expresses tests for signifiers of the class of interest as natural language prompts.
    The prompts are combined with unlabeled examples and given to a pre-trained language model. The model's responses are mapped to votes on the true label for the example.}
    \label{fig:overview}
\end{figure}

Limited labeled training data is a major bottleneck in many areas of supervised machine learning.
In recent years, the area of \emph{programmatic weak supervision}~\citep{zhang:arxiv22} has emerged to address this bottleneck.
There are a range of techniques, but generally they use multiple noisy heuristic labelers called \emph{labeling functions}, such as hand-written code and other models, to create training data for new tasks.
These labelers are applied to abundant unlabeled data, and they either vote on the correct label or abstain.
Then, a label modeling stage attempts to resolve the conflicts among the labelers without access to much or any ground truth labels.
The resulting labels are finally used to train an end model that generalizes beyond the labelers.
This approach has seen many practical successes in areas such as information extraction~\citep{callahan:npjdigmed19, ratner:vldbj20, fries:nc2021} and medical imaging~\citep{dunnmon:patterns20, eyuboglu:naturecommunications21}.
Programmatic weak supervision has also been deployed at major technology companies~\citep{bach:sigmod19-industrial, bringer:deem19,suri:vldb20,kuang:aistats22}.
Large pre-trained language models are an untapped resource as a potentially complementary source of heuristic labels.
In addition to the ease of specifying heuristics with natural language, we show that they can effectively capture a wide range of fuzzy concepts that can be hard to express as traditional labeling functions written in code.

Despite this potential, naively prompting pre-trained models to label training data has several potential pitfalls.
First, language models are sensitive to the wording of prompts~\citep{jiang:tacl20,shin:emnlp20}.
Even models that have been fine-tuned on a variety of prompt wordings can still be sensitive to phrasing~\citep{sanh:iclr22, wei:iclr22, webson:arxiv21}.
Second, prompted language models are limited in the complexity of the instructions they can follow~\citep{webson:arxiv21, mishra:acl22}.
Tasks can have nuanced decision boundaries based on context.
For example, a link to a music video might be more likely to be spam on a news website but not spam on a video site.
A single prompt, even paraphrased into multiple variants to address model sensitivity, is often insufficient to capture the full specification of a task.
For these reasons, a framework for incorporating pre-trained language models into weak supervision is needed that can incorporate significant amounts of subject matter expertise in a manner efficient for users.

Prompting is an emerging area in natural language processing, and recent related works have explored using prompted models as sources of supervision.
Several works use pre-trained models to generate or modify text examples conditioned on a desired label that can be used for training~\citep{schick:emnlp21,ye:arxiv22,wu:acl22,bonifacio:arxiv22}.
Other recent works use pre-trained models to aid in labeling unlabeled examples.
Concurrently, \citet{lang:arxiv22} use co-training to iteratively generate training data for variations of the same prompt.
Also concurrently, \citet{zhang:acl22} use prompting and labeled training data to suggest new labeling functions.
Also concurrently, \citet{chen:arxiv22} propose using embeddings from foundation models to capture which examples are best labeled by which labeling functions.
Across these methods, there remains a need for a framework that allows users to refine the contours of a decision boundary with multiple prompts, particularly when labeled data is scare.

In this work, we propose a framework for incorporating prompting into programmatic weak supervision, in order to address the above challenges and realize potential benefits from pre-trained language models (Figure~\ref{fig:overview}).
We model prompts as labeling functions by adding additional metadata that maps possible completions to target labels or abstentions.
For example, if a task is to classify spam comments, a prompt could be ``Does the following comment ask the user to click a link?''
If the language model responds positively, then this is an indication that the comment is spam.
On the other hand, if the model responds negatively then that might be mapped to an abstention because both spam and non-spam comments can lack that property.
We then model the outputs of the these labeling functions as usual: using a label model to reason about the accuracies of the different prompts and create training data for an end model.
This approach is novel because it exploits pre-trained language models not just as zero- or few-shot learners, but as rich sources of knowledge that can be queried in many complementary ways to create training data.

We conduct an extensive experimental study of this approach.
Using the WRENCH~\cite{zhang:neurips21} benchmark as a starting point, we first demonstrate that many existing types of labeling functions expressed as code can be effectively translated into natural language prompts.
We show on a range of GPT-3~\citep{brown:neurips20} and T0~\citep{sanh:iclr22} models that using these prompts for zero-shot querying and using the resulting prompted predictions as labeling functions leads to end models that are more accurate than those trained on the original labeling functions.
Surprisingly, we find that using these translated labeling functions works better in many cases than simply prompting the model to solve the task of interest.
This result suggests that pre-trained models contain more useful information than can be easily accessed by a single zero-shot prompt.
The additional domain knowledge provided by expressing complementary heuristics as prompts and describing how they relate to the task of interest is a key ingredient for improved accuracy.
We show empirically that these prompt-based labeling functions usually make complementary, i.e. only weakly correlated mistakes, suggesting that the pre-trained language is actually applying different heuristics based on different prompts.

In summary, our main contributions are:
\begin{itemize}
    \item We propose expressing wide ranges of data-labeling heuristics as zero-shot prompts for pre-trained language models, and using a label model to resolve their conflicts.
    \item We demonstrate the effectiveness of this new approach as a zero-shot learning approach, showing that prompting pre-trained models with multiple heuristic tasks can significantly outperform directly prompting the model to solve the task of interest, with an average improvement of 20.2 percentage points.
    \item We also show that translating labeling functions expressed as code into prompts can lead to significantly improved weakly supervised models, with an average improvement of 7.1 percentage points, when using our best language model, \tzeropp~\citep{sanh:iclr22}
\end{itemize}

\section{Related Work}

This work builds on both weakly supervised machine learning and prompting with large pre-trained language models.
In this section, we overview the most closely related work.

\subsection{Weakly Supervised Machine Learning}

The difficulty of obtaining large amounts of labeled training data has long motivated alternatives to traditional supervised machine learning.
\emph{Weak supervision} refers to a broad family of techniques that attempts to learn from data that is noisily or less precisely labeled than usual.
Our focus is on \emph{programmatic weak supervision}, in which the sources of supervision are heuristic labelers, often called \emph{labeling functions} that vote on the true labels of unlabeled examples~\citep{zhang:arxiv22}.
Labeling functions can be hand-written programs, models trained for related tasks, or even human annotators if available.
Labeling functions have their roots in work on \emph{distant supervision}~\citep{craven:ismb99,mintz:acl09}, in which a single heuristic is used to label data and the resulting labels are assumed to be noise-free.
\citet{ratner:neurips16} proposed the \emph{data programming} paradigm for weak supervision, in which multiple labeling functions that can disagree or abstain are available.

Using multiple labeling functions gives rise to the key technical challenge in programmatic weak supervision: resolving their disagreements without access to ground truth, in order to create training data.
The original formulation of data programming uses a probabilistic generative model that assumes the ground truth label for each example is a latent random variable that generates the outputs of the labeling functions.
The parameters of the model are learned by maximizing the likelihood of the observed outputs of the labeling functions.
This model generalizes the classic Dawid-Skene model~\citep{dawid:royalstats79} for \emph{crowdsourcing}, i.e., learning from multiple human annotators.
In the simplest case, the label sources can be assumed to be conditionally independent given the true label.
In practice, this approach often works well.
However, since programmatic heuristics might exhibit biases and correlations in more systematic ways than human annotators, it is often advantageous to model more complex dependencies among the labeling functions.
Multiple methods for learning such dependencies from the labeling function outputs have been proposed~\cite{bach:icml17,varma:icml19,ratner:aaai19}.
Many of these techniques for data programming are integrated in the Snorkel system~\citep{ratner:vldbj20}.

Programmatic weak supervision has been extended in many directions.
Using adversarial learning instead of maximum likelihood estimation can provide strong theoretical guarantees without assumptions on the distribution of labels and labeling function outputs, but requires either a small amount of labeled data or other assumptions to constrain the accuracy of the labeling functions~\citep{arachie:jmlr21, mazzetto:aistats21, mazzetto:icml21}.
Weak supervision can be applied to other settings like structured prediction~\citep{sala:neurips19,safranchik:aaai20,shin:iclr22}.
Labeling functions can incorporate additional forms of supervision beyond individual labels, such as hierarchical multi-task supervision~\citep{ratner:aaai19}, partial labels~\citep{yu:aistats22}, labels from misaligned spaces~\citep{zhang:iclr22}, or constraints~\citep{arachie:uai21}.
Labeling functions can also be automatically constructed using a small amount of labeled data~\citep{varma:vldb18}.
Another line of work has extended the label modeling stage to incorporate features of the underlying data, in order to model which types of examples each labeler is best at labeling~\citep{varma:arxiv16}.
Concurrent with our work, \citet{chen:arxiv22} proposed using large pre-trained models to create representations for the label model.
Our work differs in that we use large pre-trained models to directly implement labeling functions as zero-shot predictors.

Finally, programmatic weak supervision is complementary to many other techniques for learning with limited labeled data.
It can be combined with semi-supervised learning~\citep{karamanolakis:naacl21}, self-supervised learning~\citep{yu:naacl21}, and active learning~\citep{brust:ki20,biegel:iclrws21}.
Since our work creates labeling functions that can be modeled in the same way as traditional ones, they can also be incorporated into all of these related frameworks.

\subsection{Language Models and Prompting}

Language models are trained to predict the next or missing words conditioned on a partial sequence of natural language text.
Neural-network-based language models have become ubiquitous in recent work in natural language processing because they learn useful vector representations of text that can be incorporated into models for other tasks.
Most recently developed language models are based on transformer architectures~\citep{vaswani:neurips17}.
Recently, there has been increasing interest in \emph{prompting}, an alternative way of exploiting language models~\citep{liu:arxiv21}.
Instead of using language models only as feature encoders, prompting uses a language model's ability to predict words to directly solve tasks.
Tasks are posed as natural language text called prompts, and the language model's predictions for missing or subsequent words are interpreted as task solutions.
The language model can either be fine-tuned on specific prompts using labeled examples, or it can be queried in a zero-shot fashion, i.e., prompted to solve tasks it has never been explicitly trained to solve.

\citet{brown:neurips20} demonstrated that large pre-trained language models can solve zero-shot tasks.
Other works showed that the zero-shot abilities of large language models can be improved by further fine-tuning the language model on a large mix of prompted tasks~\citep{mishra:acl22,sanh:iclr22,wei:iclr22}.
Despite these successes, there are still many challenges when using prompting for zero-shot or few-shot learning.
Models can be sensitive to the wording of the prompt~\citep{jiang:tacl20,shin:emnlp20,sanh:iclr22, wei:iclr22, webson:arxiv21}, and many works have tried to reduce this sensitivity and boost accuracy~\citep{mishra:acl22,sanh:iclr22,wei:iclr22}.

Several recent works have investigated other ways of creating or augmenting supervision using pre-trained language models.
\citet{schick:emnlp21} prompt language models to generate examples of a certain label, e.g., generating documents with a specific topic.
\citet{ye:arxiv22} generate data in an unsupervised way and then label them for training using a simple classification rule.
\citet{chia:acl22} generate examples expressing relations among entities to create training data for relation extraction.
\citet{wu:acl22} fine-tune language models to modify datasets so that they exhibit fewer biases, and \citet{bonifacio:arxiv22} fine-tune them to modify datasets for different information retrieval tasks.
Several works use language models to generate ``chains of thought'' that can improve reasoning and be used for self-training~\citep{wei:arxiv22,wang:arxiv22,zelikman:arxiv22}.
In concurrent work, \citet{lang:arxiv22} use co-training to fine-tune language models, where the different views of the data come via different prompts.
Like other work on enforcing consistency among prompted outputs~\citep{elazar:tacl21,anonymous:arr22}, they consider alternative wordings of the same task, whereas we focus on prompting multiple tasks to create supervision.
Also in concurrent work, PRBoost~\citep{zhang:acl22} uses labeled data and labeling function templates to prompt language models to suggest additional labeling functions to human annotators.
In contrast, we show that no modification of existing weak supervision pipelines are needed to achieve good performance, and that sufficiently large pre-trained language models are powerful sources of weak supervision.

\section{Weak Supervision via Prompting}

In this section we describe our proposed approach to incorporating large pre-trained language models into weakly supervised machine learning.
The goal is to enable data scientists and other subject matter experts to leverage these resources more effectively.
We focus on scenarios where users are not necessarily machine learning experts, meaning that fine-tuning large models with gradient updates is either infeasible because of the size of the model or impossible because they do not have access to the underlying model.
Instead, they might only have API access and want to exploit the large pre-trained model to create a new one that is higher quality and servable in production (i.e., not prohibitively large to work with).
Our presentation and experiments in Section~\ref{sec:results} focus on the case where all supervision comes via a language model, but this approach also naturally integrates with other forms of weak supervision, such as hand-engineered programs.

\subsection{Workflow}

\begin{figure}[t]
    \centering
    \includegraphics[width=1.0\textwidth]{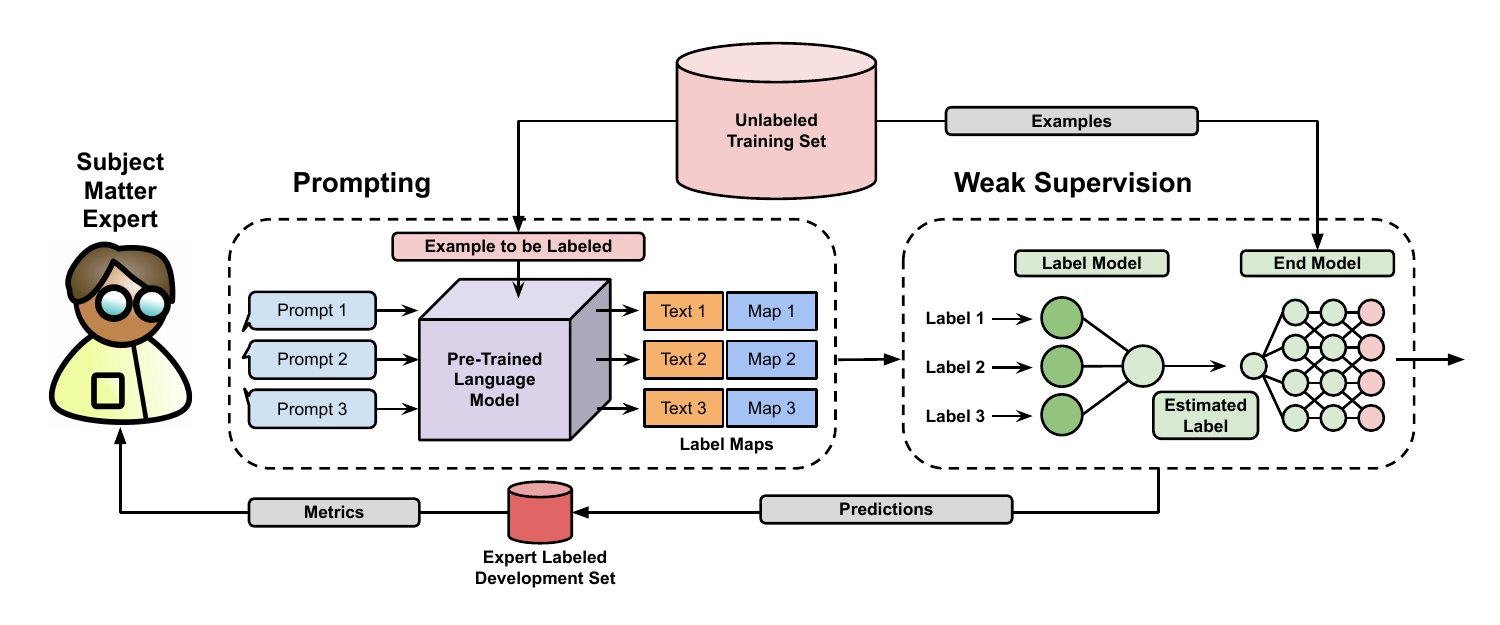}
    \caption{Language models in the loop: the overall framework for developing and applying prompted labeling functions. The subject matter expert (SME) expresses their domain knowledge via prompts that are combined with unlabeled examples and given to a pre-trained language model. The model's responses are interpreted with label maps to produce votes on the true label. These votes are denoised with a label model, and the resulting estimated labels are used to train an end model. Throughout the process, the SME can refine their prompts by inspecting unlabeled examples and evaluating with a small labeled development~set.}
    \label{fig:system}
\end{figure}

We first describe the workflow in our approach (Figure~\ref{fig:system}).
In the the scenarios we consider, the user is a subject matter expert (SME) who wants to create a classifier for unlabeled data.
Continuing our running example, this could be a classifier for detecting spam comments on a video website.
They have access to a large amount of unlabeled data that can be used for training.
They also have access to a small (dozens up to hundreds of examples) development data set that has been manually labeled.
That development set will be used to evaluate modeling decisions like the choice of prompts and tuning hyperparameters.

The SME then develops heuristics for labeling examples by inspecting unlabeled examples.
These heuristics are expressed as natural language prompts that capture some aspect or feature of the data that is likely to indicate the true label.
For example, in the case of labeling spam comments, the SME might notice by browsing comments that many spam examples contain some call to action, such as asking the reader to click or visit a link.
Enumerating all the ways that a call to action could be expressed in natural language is challenging to do accurately, requiring the SME to curate many keywords and regular expressions that are sufficiently precise.
Alternatively, a simple prompt like ``Does the following comment ask the reader to do something?'' has the potential to better capture this heuristic while requiring less effort from the SME.

The SME's heuristic prompts are encapsulated as \emph{prompted labeling functions}.
Prompted labeling functions consist of a prompt template and a label map.
The prompt template defines how the SME's prompt is applied to unlabeled examples.
Unlabeled examples consist of one or more fields of text.
In this work, we focus on Yes/No question answering-style prompt templates.
However our method generalizes to many prompt template and label map formats.
In the case of website comments, the text could be represented as a single field \texttt{[TEXT]} and the entire prompt template for a labeling function could be
\[
\texttt{Does the following comment ask the reader to do something? [TEXT]}
\]
The label map then defines how responses by the pre-trained language model are mapped to votes on the true label for the example.
Our framework focuses on generative language models like T0~\cite{sanh:iclr22} and GPT-3~\cite{brown:neurips20}, so the responses can be arbitrary text strings.
The label map $M: \mathcal{S} \rightarrow \mathcal{Y} \cup \{\emptyset\}$ is a function from the set $\mathcal{S}$ of strings composed from the pre-trained language model's vocabulary to the set of labels $\mathcal{Y}$ and a special symbol $\emptyset$, which indicates that the labeling function abstains, i.e., has no vote on that example.
In the case of the above example prompt, a corresponding label map would map positive responses like ``Yes'' and ``True`` to the spam label, and all other responses to abstentions.
SMEs can also refine their prompts by evaluating their labeling functions on the unlabeled data and the small labeled development data set.
In this way, the SME enters a feedback loop, in which they can reword prompts and construct additional ones to add complementary information.
We discuss the development of prompted labeling functions further in Section~\ref{sec:developing}.

After the SME has developed their prompted labeling functions, they can be plugged into many standard weak supervision frameworks, such as Snorkel~\cite{ratner:vldbj20}.
In such frameworks, the labeling functions are executed on all the available unlabeled data to produce votes on what the correct label is.
These votes are aggregated in the \emph{label model} that produces probabilistic estimates of the correct label.
Finally, an appropriate end model, such as a deep neural network, is trained for the classification task of interest by minimizing the expected empirical risk with respect to the probabilistic estimates of the true labels.
The resulting classifier can be used outside of this weak supervision framework and independently from the underlying pre-trained language model.
In this way, language models in the loop enable SMEs to distill information locked away in large foundation models into smaller, more servable models.
As we show in Section~\ref{sec:results}, these resulting models can also often significantly improve over the accuracy obtained by using the pre-trained model alone.

\subsection{Developing Prompted Labeling Functions}
\label{sec:developing}

We now discuss the advantages of writing prompted labeling functions, and how it differs from writing labeling functions in code.
Prompted labeling functions are a mechanism by which a large pre-trained model can be adapted with limited labeled training data to new tasks.
We find that large pre-trained models such as \tzeropp~and GPT-3 exhibit a phenomenon wherein they ``know more than they realize,'' in the sense that they can solve many other tasks that provide useful signals about the task of interest, even if they do not know how to integrate those signals.

Weakly supervised machine learning is a natural paradigm for integrating these signals effectively.
For example, in the spam comment task, the zero-shot approach is to prompt the pre-trained language model with a prompt like ``Is the following comment spam?''
In contrast, we propose using prompting to collect multiple signals related to the task of interest.
Examples from our experimental study (Section~\ref{sec:results}) are
\begin{enumerate}
    \item ``Does the following comment ask the reader to do something?''
    \item ``Does the following comment reference the speaker's channel?''
    \item ``Does the following comment contain the words `check out'?''
\end{enumerate}
Each of these prompts, along with the associated label map, provides additional domain knowledge about the definition of spam in this particular application.
Task supervision is often multifaceted and difficult to summarize in a single prompt.
Pre-trained language models can have difficulty with long, nuanced instructions~\citep{mishra:acl22,webson:arxiv21}.
Our approach breaks down task supervision into salient components, expressed as multiple prompts capturing different aspects of labeling.

The above example prompts also illustrate the advantages that pre-trained language models can offer weakly supervised machine learning.
Standard rule-based labeling function expressed in code or via resources like term dictionaries are brittle.
In contrast, prompts can handle significant amounts of ambiguity.
The three example prompts above are arranged in order of decreasing ambiguity. 
Prompt (1) covers a wide range of scenarios that would be difficult to enumerate with rules.
Answering the prompt accurately likely requires an understanding of intent.
Prompt (2) is in the middle, in that it asks for references to a specific entity (the speaker's channel), but that entity can be referred to in many ways, including indirectly, e.g., a comment like ``Like and subscribe!''
Prompt (3) is the most specific, asking if the comment contains a specific phrase.

\begin{figure}[t]
    \centering
    \includegraphics[width=1.0\textwidth]{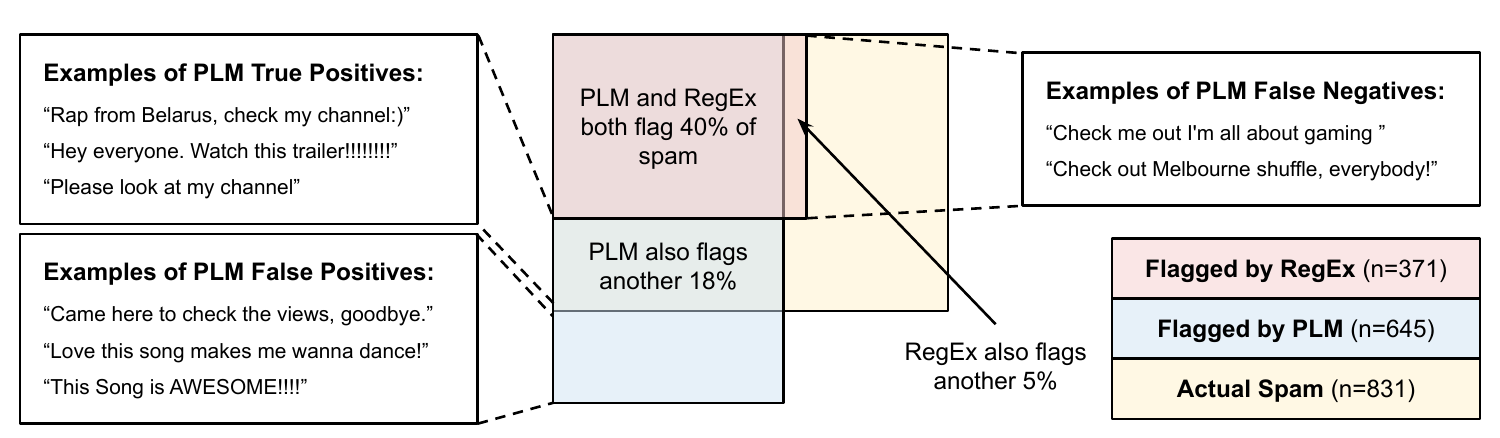}
    \caption{A comparison of a regular expression (RegEx) labeling function from the WRENCH benchmark~\citep{zhang:neurips21} with the corresponding prompted labeling function using the \tzeropp~\citep{sanh:iclr22} pre-trained language model (PLM). The regular expression looks for variations of the phrase ``check out'' and the prompted labeling function uses the prompt ``Does the following comment contain the words `check out'?'' RegEx has 100\% precision and 45\% recall, while PLM has 76\% precision and 58\% recall. This comparison shows that even simple labeling functions can be made more general while maintaining acceptable precision by using prompting.}
    \label{fig:lf_breakdown}
\end{figure}

Surprisingly, even prompted labeling functions asking for a specific phrase have interesting, useful properties that differ from traditional labeling functions.
Figure~\ref{fig:lf_breakdown} compares a prompted labeling function using prompt (3) with the corresponding,  traditional labeling function from the WRENCH benchmark for weak supervision~\citep{zhang:neurips21} on the Youtube comment spam dataset.
The traditional labeling function is a regular expression that also checks for the phrase ``check out.''
It is very precise, with 100\% precision and 45\% recall.
The prompted labeling function has 76\% precision and 58\% recall.
The tradeoff is that the prompted labeling function finds many true positives that say something with a meaning similar to ``check out,'' but also misfires on some false positives.
This example illustrates that even with seemingly straightforward heuristics like a simple regular expression, pre-trained language models can provide useful additional flexibility.
Our experiments in Section~\ref{sec:results} show that this can be a favorable tradeoff for developers.

\subsection{Calibration}
We find that it is useful to improve the calibration of prompted labeling functions.
Calibration is a measurement of how strongly a model's predicted probabilities correlate with observed accuracy, i.e., a predicted probability of $\hat{p}$ should be correct $\hat{p} \cdot 100$\% of the time. 
Current language models are not well-calibrated, with predicted probabilities subject to several forms of biasing, e.g., favoring tokens observed more during pretraining or tokens that appear near the end of a prompt \cite{jiang:tacl20,zhao:icml21}.
Miscalibration creates challenges in prompting, which requires choosing the most likely answer from a set of candidate text completions. 
When using prompts as labelers, we may also want to threshold predictions to select only the most confident answers.
Popular recalibration methods such as Platt and vector scaling \cite{platt:book99, guo:icml17} require labeled data to learn a transformation of the model's predicted probabilities, creating challenges to directly applying these methods in zero-shot settings.
Instead, we use \textit{contextual calibration} \cite{zhao:icml21}, where scaling weights are estimated from the predicted token probabilities of a prompt queried using ``content-free" or null input instances.
Contextual calibration has demonstrated empirical performance gains when used in prompt-based, few-shot classification.  
We use the tokens \{ \texttt{N/A}, $\epsilon$, \texttt{[MASK]}, \texttt{NULL}, \texttt{<|endoftext|>} \} as our null inputs, using the average predicted probabilities per token to estimate our scaling weights for each prompt. 
The resulting transformation is then applied to each prompted labeling function's predictions.

\section{Experimental Study}
\label{sec:results}

We conduct an experimental study to evaluate how incorporating prompted labeling functions compare with two alternatives: (1) distilling pre-trained language models in a zero-shot fashion, and (2) hand-written labeling functions.
We use the WRENCH benchmark~\citep{zhang:neurips21} for weak supervision in order to control the choice of labeling functions.
WRENCH provides traditional labeling functions that we translate into corresponding prompted labeling functions for comparison.
We find that
\begin{enumerate}
    \item Creating models via prompted labeling functions can significantly outperform directly prompting the model to solve the task of interest, with an average improvement of 20.2 percentage points, and
    \item Translating labeling functions expressed as code into prompts can lead to significantly improved weakly supervised models, with an average improvement of 7.1 percentage points, when using our best language model, \tzeropp~\citep{sanh:iclr22}.
\end{enumerate}

\subsection{Datasets}
The WRENCH benchmark includes 22 diverse datasets for evaluating weakly supervised learning \cite{zhang:neurips21}. 
Datasets include labeling function sets for programmatically creating labeled training data and corresponding manually curated gold labels for evaluation.  
We focus on a subset of text classification tasks: YouTube, SMS, and Spouse.
Note that 4 WRENCH datasets (IMDB, Yelp, AG News, TREC) were used as part of \tzeropp~training, thus we exclude them from our analysis.
Dataset summary statistics are outlined in Table~\ref{tab:datasets}.

%
%
\begin{table}[!ht]
\centering
\begin{tabular}{l|l|c|l|l|c|rrr} \toprule
Name    & Task & \#Labels & Class Labels & $P$(positive) & \#LFs & Train & Valid. & Test  \\ \midrule
YouTube & Spam Detection      & 2   & \texttt{HAM,SPAM}      & 0.488 (0.02)   & 10    & 1,586  & 120        & 250   \\
SMS     & Spam Detection      & 2   & \texttt{HAM,SPAM}        & 0.132 ($<$0.01)   & 73    & 4,571  & 500        & 500   \\
Spouse  & Relation Extraction & 2   & \texttt{NOT\_SPOUSE,SPOUSE}  & 0.074 ($<$0.01)   & 9     & 22,254 & 2,801      & 2,701 \\ \bottomrule
\end{tabular} 
\caption{Summary statistics for our WRENCH text classification datasets. $P$(positive) is the class balance \textbf{}of the positive label (\texttt{SPAM} or \texttt{SPOUSE} depending on the task) calculated as the mean and standard error of relative frequency for all gold labeled splits.} 
\label{tab:datasets}
\end{table}

\subsection{Translating WRENCH Labeling Functions into Prompts}
Labeling functions are developed by SMEs via data exploration, which entails iteratively designing labeling rules by inspecting unlabeled examples and a small, hand-labeled development set.
For WRENCH datasets, this process has already occurred, so our experiments focus on translating existing labeling rules into prompted form.
We note this is a more restricted setting than if SMEs developed prompts initially, as WRENCH labeling functions are biased towards rules that are easy to express in code while prompts have more flexibility.  
All labeling function prompts are formulated as Yes/No questions and a label map that transforms text completions into class labels or abstains (i.e., not emitting a label).   

For example, consider a WRENCH labeling function written in Python for the Spouse task, which uses keywords occurring between person mentions to label negative training examples by identifying likely family members. 
\begin{verbatim}
def lf_familial_relationship(x):
    family = {"father", "mother", "sister", "brother", "son", "daughter", "uncle", "aunt"}
    return NOT_SPOUSE if len(family.intersection(set(x.between_tokens))) > 0 else ABSTAIN}
\end{verbatim}
Instead of enumerating an incomplete list of keywords describing family relationships, our prompt focuses on the general insight conveyed by the labeling function. 
\begin{align*}
&\texttt{Context: [TEXT]\textbackslash n\textbackslash nAre [PERSON1] and [PERSON2] family members?} \\ 
& \mapsto \texttt{\{yes:NOT\_SPOUSE, no:ABSTAIN\} }
\end{align*}
Prompts were developed for GPT-3 and \tzeropp~separately by iteratively querying each language model with unlabeled training instances, performing an ad hoc performance assessment, and then selecting a single prompt to use per labeling function.
This mirrors the process by which a SME might query a language model to guide prompt development.
The complete list of WRENCH prompts used in this work are found in Appendix \textsection \ref{section:wrench_lf_prompts}.

\subsection{Comparing Programmatic Labelers}

%
%
\begin{table}[!ht]
\centering
\begin{tabular}{lll}
\toprule
Dataset & Model & Prompt \\
\midrule
YouTube & \tzeropp & \texttt{Is the following comment spam?\textbackslash n\textbackslash n"[TEXT]"}  \\
SMS & \tzeropp & \texttt{Is the following text message spam?\textbackslash n\textbackslash n"[TEXT]"} \\
Spouse & \tzeropp & \texttt{Context: "[TEXT]"\textbackslash n\textbackslash nAre [PERSON2] and [PERSON1] married?} \\ \midrule
YouTube & GPT-3 & \texttt{Q: Is the following comment "[TEXT]" spam? \textbackslash nA:} \\
SMS & GPT-3 & \texttt{Q: Is the following text message "[TEXT]" spam? \textbackslash nA:} \\
Spouse & GPT-3 & \texttt{Context: "[TEXT]"\textbackslash nQ: Are [PERSON1] and [PERSON2] married? \textbackslash nA:} \\
\bottomrule
\end{tabular}
\caption{Zero-shot prompts for all datasets and language model families.
\texttt{[TEXT]}, \texttt{[PERSON1]}, \texttt{[PERSON2]} are populated with text from the target example.
Label maps are \texttt{\{no:HAM, yes:SPAM\}} for YouTube/SMS and \texttt{\{no:NOT\_SPOUSE, yes:SPOUSE\}} for Spouse. }
\label{tab:zeroshot}
\end{table}

We compare three approaches for programmatically generating training labels, following the typical workflow used for weakly supervised learning. 
For each dataset in our analysis, we assume the original training split is unlabeled.
All \emph{labelers}, here prompted labeling functions and code-based labeling functions, are applied to the unlabeled training split to generate votes for the true label of each example.
All prompts are calibrated using contextual calibration.
All labeler votes, unless otherwise noted, are combined and denoised using the FlyingSquid \cite{fu:icml2020} label model to estimate a single, probabilistic consensus label per example. 
The resulting labels are used to train a RoBERTa \cite{liu:openreview2020} end model, which provides a smaller, more servable classification model tailored to our task of interest.
All model performance measures are then evaluated using gold labeled test splits. 
The three approaches we compare are:
\begin{enumerate}
\item \emph{WRENCH Benchmark}: The original WRENCH labeling functions released as part of the benchmark.
Here \textit{majority vote} (i.e., the mode of all labeling function outputs per example) is used as the label model since it performed the best when used with RoBERTa for all three of our tasks.

\item \emph{Zero Shot}: A zero-shot baseline where training data is labeled by one prompt that queries a language model for an example's class label. 
Prompts are outlined in Table \ref{tab:zeroshot} and were designed to align with prompts commonly used in zero shot learning by providing a simple, but often underconstrained, task description.

\item \emph{Prompted Weak Supervision}: The prompted versions of the WRENCH labeling functions.
These labelers reflect the prototypical weakly supervised workflow, except we have replaced manually coded labeling functions with prompted versions.
\end{enumerate}

\subsection{Large Language Models}
All prompts are evaluated using two different language model families: GPT-3 and \tzeropp.
We use the InstructGPT \cite{ouyang:corr22} family of GPT-3 engines, evaluating Ada, Babbage, and Curie since different engines are claimed to be better suited to specific tasks.\footnote{See https://beta.openai.com/docs/engines/gpt-3}. 
DaVinci was not used due to cost constraints (see complete pricing for all GPT-3 queries in  Appendix \S \ref{sec:cost}). 
All queries were submitted via the OpenAI API between 01/24/2022--03/01/2022.
Queries were restricted by the API to include only the top 100 most likely text completions. 

\tzeropp~\citep{sanh:iclr22} is an open, publicly available 11B parameter model based on the T5 architecture \cite{raffel:jmlr20}.
\tzeropp~is trained using a large dataset of supervised tasks transformed into prompted training data. 
This explicit, multitask formulation of prompted training data results in better zero-shot classification performance that often matches or exceeds the much larger GPT-3.
The model requires 42 GB of GPU memory to efficiently run locally without parameter offloading.
We used a p3.8xlarge AWS EC2 instance with 4 Tesla V100 GPUs for inference.

\subsection{Evaluation Metrics}
We evaluate all models using precision, recall, F1, and accuracy. 
Performance metrics are reported as the mean and standard error of six training runs using different random seeds.
Standard error is calculated using the sample standard deviation.
For direct comparisons with WRENCH, we report accuracy or F1 based on the default metric reported in WRENCH benchmarks.

\subsection{Results}

%
%
\begin{table}[!ht]
\centering
\begin{tabular}{ l cc| cc | cc }
\toprule
& \multicolumn{2}{c}{Youtube (Accuracy)} & \multicolumn{2}{c}{SMS (F1)} & \multicolumn{2}{c}{Spouse (F1)} \\ \cmidrule(lr){2-3}\cmidrule(lr){4-5}\cmidrule(lr){6-7}
                    & Zero Shot  & Prompted WS          & Zero Shot  & Prompted WS         & Zero Shot   & Prompted WS           \\ \midrule
WRENCH Benchmark    & -          & 94.9 (0.5)           & -          & 92.4 (0.5)          & -           & 37.9 (2.8)            \\ \midrule
\tzeropp                & 58.7 (2.4) & \textbf{92.0 (0.5) } & 83.2 (2.4) & \textbf{91.8 (1.6)} & 41.5 (13.1) & \textbf{ 62.9 (0.8) } \\
InstructGPT Curie   & 52.8 (0.0) & 77.7 (1.9)           & 0.0 (0.0)  & 65.7 (5.8)          & 49.6 (1.0)  & 41.0 (0.9)            \\
InstructGPT Babbage & 78.5 (3.0) & 85.1 (1.3)           & 32.2 (3.0) & 23.6 (0.0)          & 40.9 (0.9)  & 34.9 (1.7)            \\
InstructGPT Ada     & 51.7 (2.4) & 52.9 (0.1)           & 26.3 (2.6) & 28.3 (1.8)          & 19.1 (0.8)  & 17.7 (6.2)            \\
\bottomrule
\end{tabular}
\caption{Performance metrics for Zero Shot and Prompted Weak Supervision (Prompted WS) using four large language models and calibrated prompts.
Scores are the mean/standard error of 6 training replicates with the best prompted model performance in bold.}
\label{tab:prompt_results_main}
\end{table}

\subsubsection{Prompted Weak Supervision}

Table \ref{tab:prompt_results_main} outlines the performance of Zero Shot and Prompted Weak Supervision using four language models (\tzeropp, InstructGPT family) compared against the WRENCH benchmark.
Prompted weak supervision outperforms the zero-shot baseline by an average of 18.2\% (-26.7 to 100\%) across all language models and datasets.
\tzeropp~consistently demonstrated strong performance, outperforming InstructGPT in all datasets when using Prompted Weak Supervision. 
Considering only \tzeropp~performance, Prompted Weak Supervision outperforms Zero Shot by an average of 39.5\% (10.3 to 56.7\%).
In the InstructGPT models, Prompted Weak Supervision largely negatively impacted performance, with performance gains consistently observed only in the YouTube dataset. 
Overall, the InstructGPT family performed substantially worse than \tzeropp, which outperformed InstructGPT Curie by an average of 37.2\% (18.4 to 53.4\%). 

Using the \tzeropp~model, prompted performance approaches or exceeds models trained using the WRENCH Benchmark labeling functions.
In the case of Spouse, \tzeropp~significantly outperformed WRENCH labeling functions, improving performance by 25 F1-score points when using Prompted Weak Supervision. 

\subsubsection{Prompt Calibration}

%
%
\begin{table}[!ht]
\centering
\begin{tabular}{ll |crrrclc}
\toprule
Dataset                  & Language Model                     & CC         & Precision            & Recall               & F1                  & $\pm$F1     & Acc.                & $\pm$Acc.   \\ \midrule
\multirow{2}{*}{YouTube} & \multirow{2}{*}{\tzeropp}              & \checkmark &  92.6 (0.5) &  91.7 (0.5) &  91.9 (0.5) &   -3.5 (0.6) & 92.0 (0.5) &  -3.4 (0.6) \\
 &               &           &  95.7 (0.4) &  95.2 (0.5) &  95.4 (0.4) &   \textendash & 95.4 (0.4) &  \textendash \\ \midrule
\multirow{2}{*}{SMS}     & \multirow{2}{*}{\tzeropp} & \checkmark &  95.9 (2.5) &  88.1 (1.1) &  91.8 (1.6) &  \textbf{ +0.3 (2.5)} & 97.9 (0.4) &  +0.2 (0.7) \\
    &              &           &  91.6 (3.2) &  91.5 (0.8) &  91.4 (1.6) &   \textendash & 97.7 (0.5) &  \textendash \\ \midrule
\multirow{2}{*}{Spouse} &  \multirow{2}{*}{\tzeropp} & \checkmark &  54.2 (1.8) &  75.4 (1.2) &  62.9 (0.8) &  \textbf{+18.0 (1.7)} & 92.8 (0.3) & \textbf{+10.0 (1.4) }\\
 &     &           &  30.7 (1.6) &  86.0 (2.8) &  44.9 (1.3) &   \textendash & 82.8 (1.2) &  \textendash \\ \midrule
\multirow{2}{*}{YouTube} & \multirow{2}{*}{InstructGPT Curie} & \checkmark &  80.1 (1.0) &  77.1 (2.1) &  76.7 (2.3) &   \textbf{+0.8 (2.0)} & 77.7 (1.9) &  -0.1 (1.6) \\
&   &           &  84.8 (0.6) &  76.4 (1.2) &  75.9 (1.3) &   \textendash & 77.7 (1.1) &  \textendash \\ \midrule
\multirow{2}{*}{SMS} &   \multirow{2}{*}{InstructGPT Curie} & \checkmark & 60.6 (11.5) &  83.8 (4.4) &  65.7 (5.8) &  \textbf{+65.7 (5.8)} & 86.2 (3.9) &  -0.4 (3.9) \\
&   &           &   0.0 (0.0) &   0.0 (0.0) &   0.0 (0.0) &   \textendash & 86.6 (0.0) &  \textendash \\ \midrule
\multirow{2}{*}{Spouse} & \multirow{2}{*}{InstructGPT Curie} & \checkmark &  29.5 (0.9) &  67.5 (2.0) &  41.0 (0.9) &  \textbf{+41.0 (0.9)} & 84.3 (0.7) &  -7.7 (0.7) \\
  &  &           &   0.0 (0.0) &   0.0 (0.0) &   0.0 (0.0) &   \textendash & 91.9 (0.0) &  \textendash \\
\bottomrule
\end{tabular}
\caption{The impact of contextual calibration (CC) on performance metrics for \tzeropp~and InstructGPT Curie, the best performing GPT-3 model when using calibrated prompts. Scores are the mean/standard error of 6 training replicates. Overall improvements due to calibration are in bold. }
\label{tab:calibration}
\end{table}

\begin{figure}[!ht]
\centering
\includegraphics[width=1.0\textwidth]{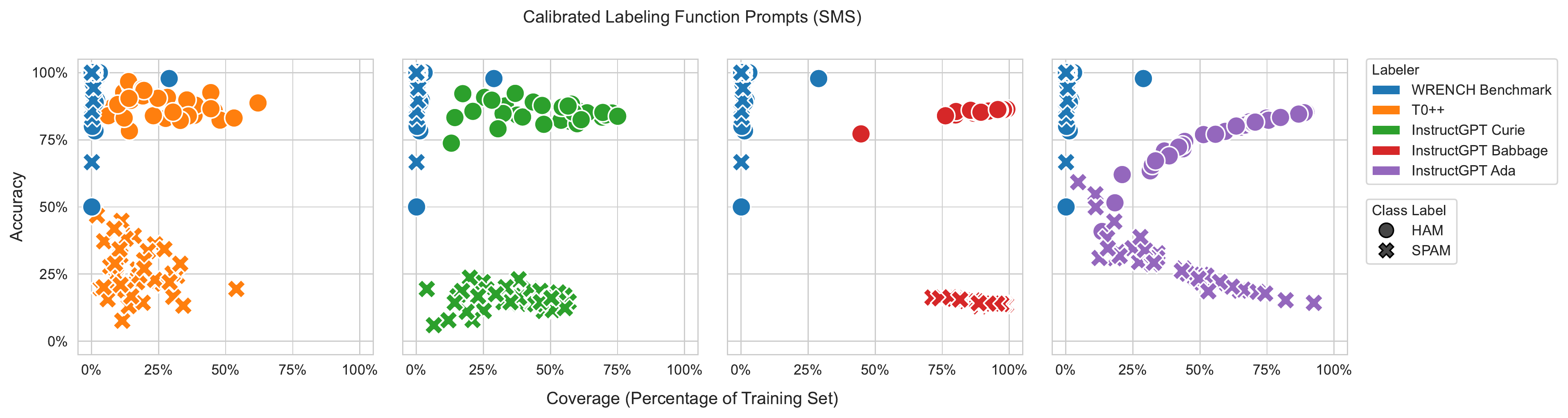}
\includegraphics[width=1.0\textwidth]{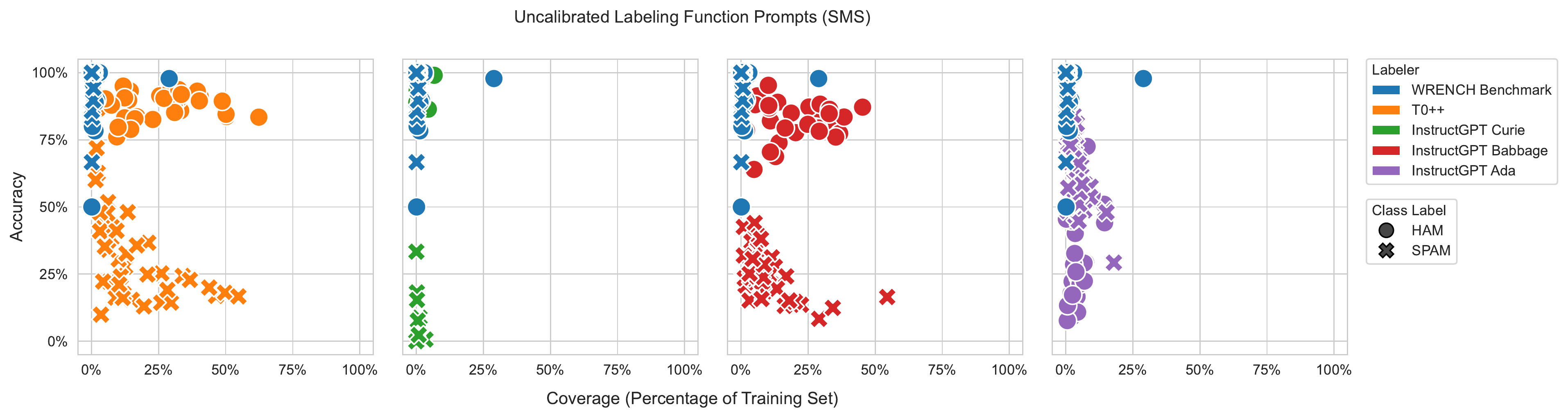}
\caption{SMS prompted labeling function coverage (x-axis) vs. accuracy (y-axis). The top figure is calibrated using contextual calibration and the bottom is uncalibrated. 
WRENCH Benchmark labeling function performance is in blue in every subfigure, which in SMS favors high precision, extremely low-coverage ($<2\%$). }
\label{fig:sms_lf_acc_cov}
\end{figure}

Calibration had significant performance impact on all language models.
Table \ref{tab:calibration} contains the overall benefit, in F1-score and  accuracy, from using contextual calibration for \tzeropp~and InstructGPT Curie.  
Complete pre- and post-calibration performance scores for all models are reported in the Appendix \S \ref{app:calibration}.
In many cases, calibration provides significant performance improvements, with the largest increases seen in cases where the uncalibrated model had pathological performance.
Figure \ref{fig:sms_lf_acc_cov} provides additional insight into calibration, where prompts evaluated with InstructGPT Curie and Ada often resulted in zero or extremely low coverage, causing training failures.
Comparing coverage and accuracy of the original WRENCH labeling functions against their prompted versions shows how prompts result in much higher coverage than the same rule as expressed in code.
For SMS, WRENCH keyword labeling functions (the blue points) are high precision, low coverage and highly tailored to the SMS task. 
Despite this low coverage, an end model trained with data generated by these labeling functions performs quite well, with 92.4 F1.
For \tzeropp~models, prompts are noisier, with higher coverage and lower accuracy especially in the positive class.
Despite this, by combining and denoising signal across multiple prompts, \tzeropp~achieves end model scores of 91.8 F1, only a 0.6 point~drop. 

\begin{figure}[!ht]
\centering
\includegraphics[width=0.87\textwidth]{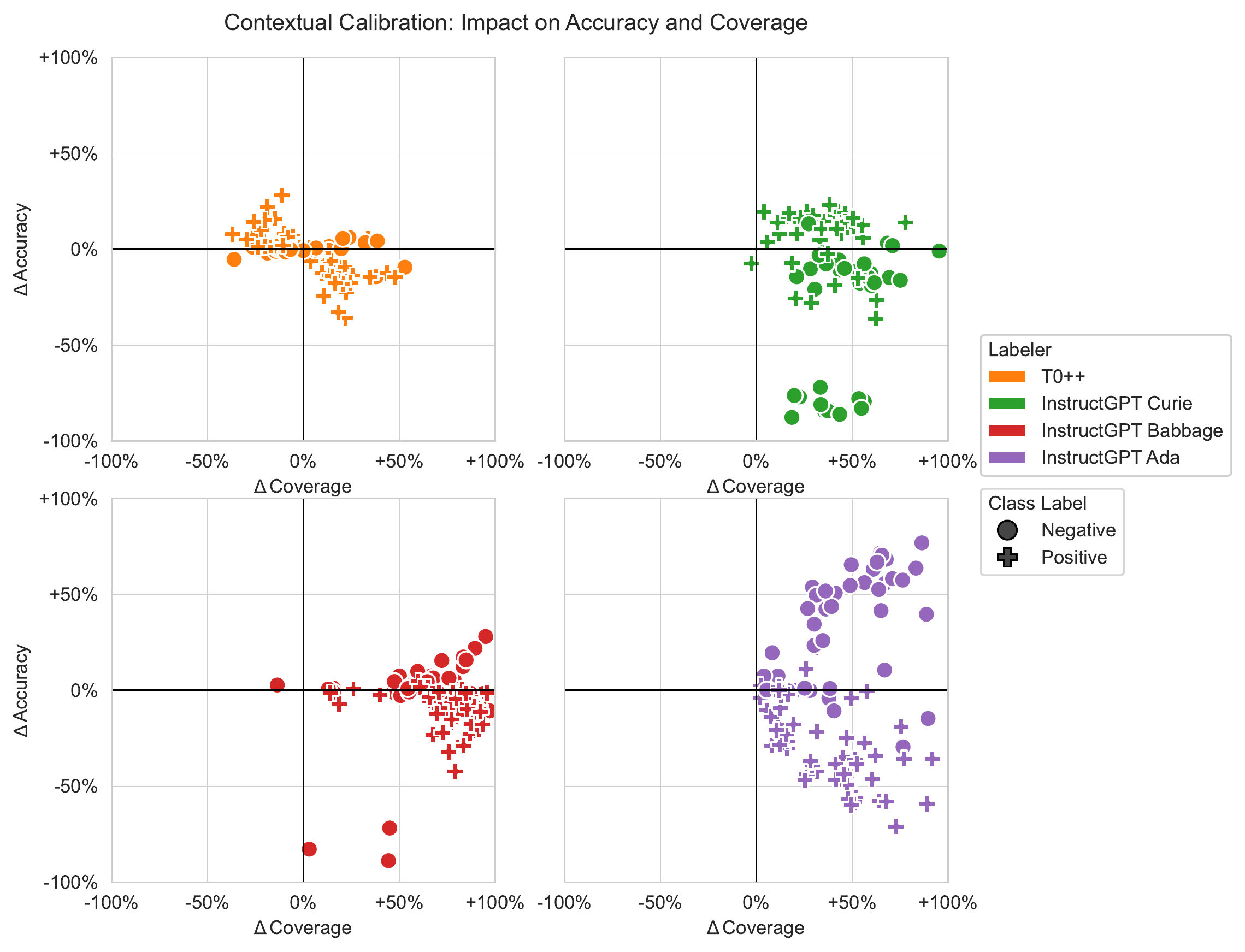}
\caption{Absolute change in accuracy and coverage after contextual calibration for all prompted labeling functions and language models. Each subfigure contains points from all datasets. The x-axis is change in coverage, the y-axis is change in accuracy, and each point reflects the change in that prompt's labeling performance after calibration.
}
\label{fig:calibration_deltas}
\end{figure}

Figures \ref{fig:calibration_deltas} shows how contextual calibration, at the level of individual prompts, can result in an unclear trade-off between accuracy and coverage.
This plot presents the absolute change in accuracy and coverage between an uncalibrated prompt its calibrated equivalent.
Recalibration generally increases a prompt's coverage, i.e., the number of labeled points, often at the cost of decreased accuracy. 
For \tzeropp~models, accuracy decreased an average of 1.5 points while coverage increased by 2.4 points.
For the InstructGPT models, the change is more substantial, with decreases in accuracy of 2.0 to 10.5 points while coverage increased by 40 to 69.7 points.
For the Babbage and Ada engines, many prompts are driven to nearly 100\% coverage, i.e., labeling the entire training set, due in part to a prompt responding with the same answer for every example.
Only \tzeropp~and InstructGPT Curie consistently improve prompt accuracy in the positive (minority) class.
The negative class in \tzeropp~had very little change in accuracy, with calibration increasing coverage at little-to-no change in accuracy.
\tzeropp~is the only language model where calibration consistently resulted in more conservative labelers, i.e., prompts where accuracy increased and coverage decreased.
Class-conditional views of these figures are available in the Appendix \S\ref{app:calibration}.

\begin{figure}[!ht]
\centering
\includegraphics[width=0.83\textwidth]{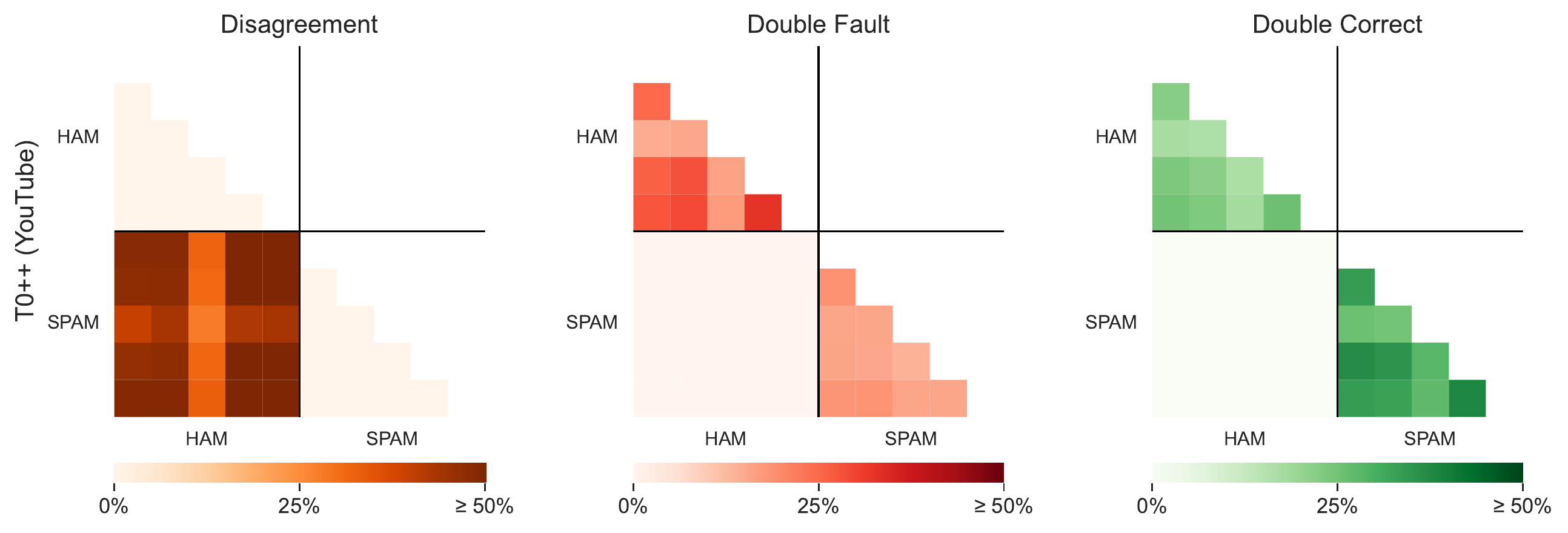}
\includegraphics[width=0.83\textwidth]{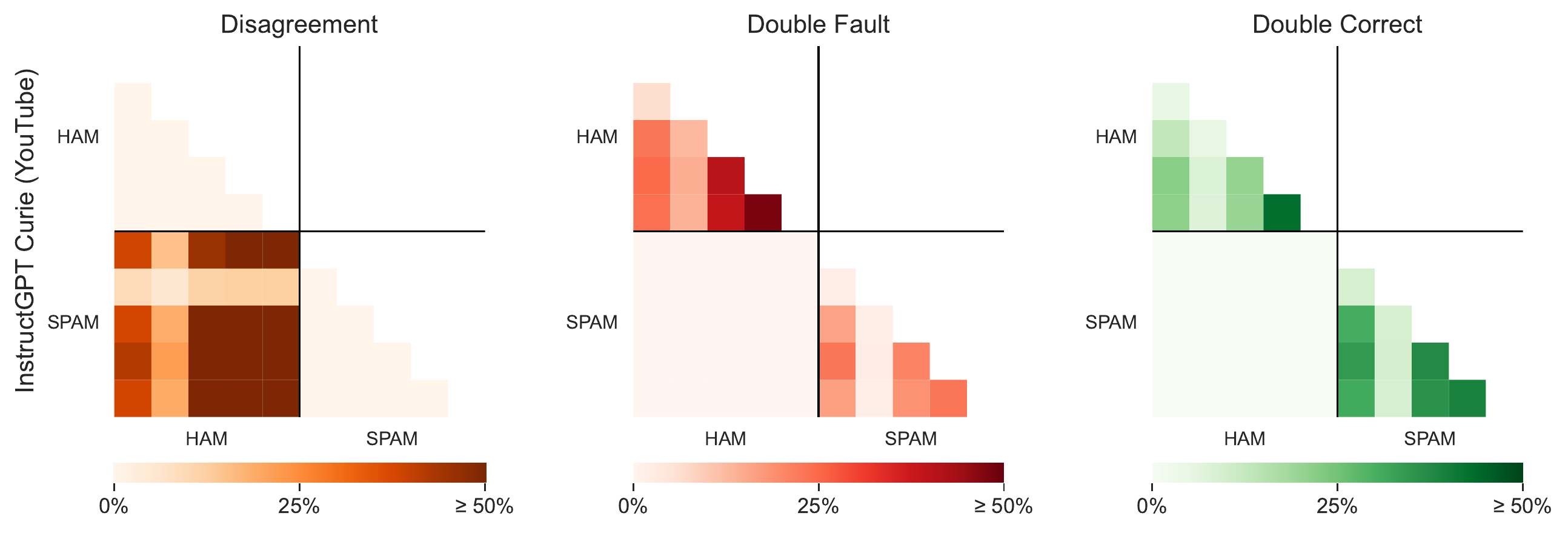}
\includegraphics[width=0.83\textwidth]{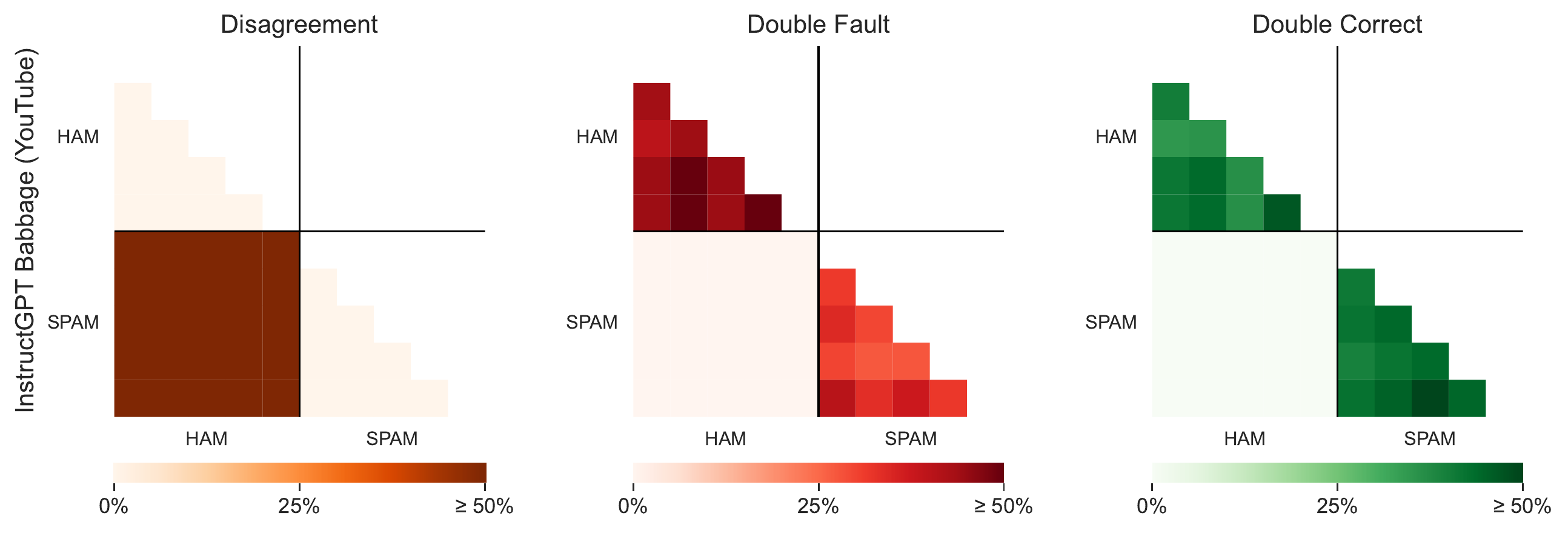}
\includegraphics[width=0.83\textwidth]{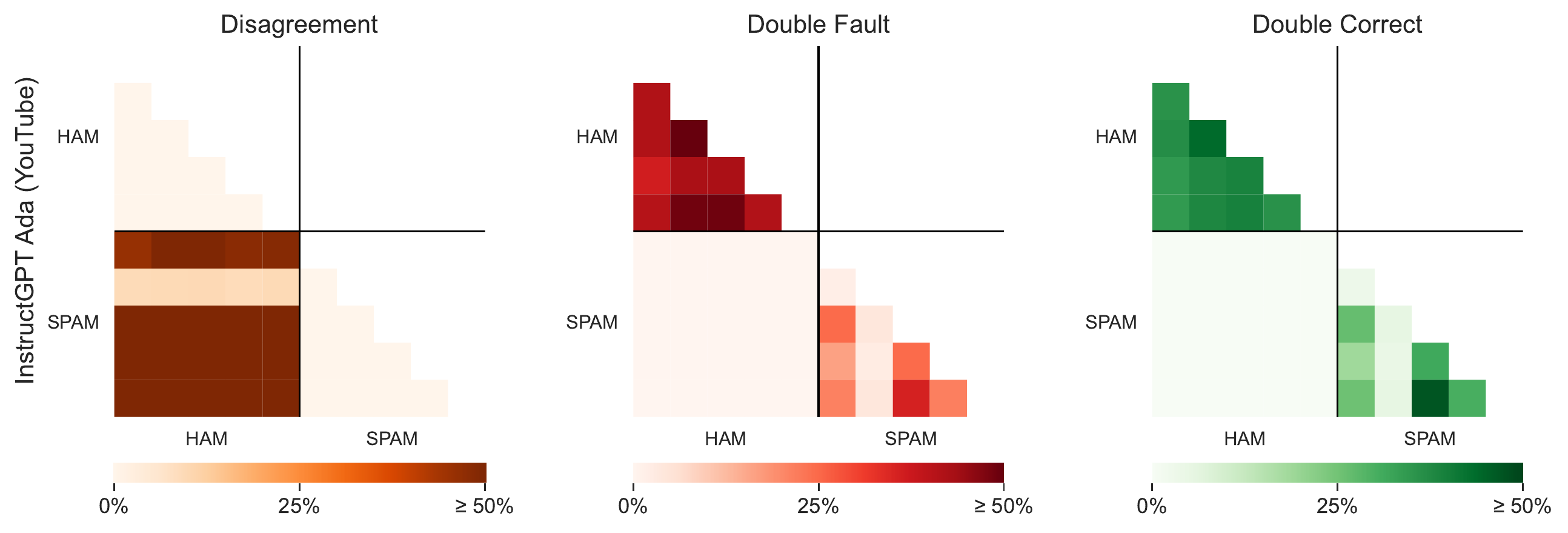}
\caption{YouTube prompted labeling function pairwise diversity measures: \textit{disagreement} (left), \textit{double fault} (center), \textit{double correct} (right). Each matrix cell represents the percentage of training examples, indicated by color intensity, where prompts $i,j$ both label an example. Rows are sorted by class label (one per-prompt) to emphasize block structure. Note some blocks are zero by definition, e.g., double fault measures when two prompts both emit the same incorrect label so the \texttt{SPAM/HAM} block is zero.    }
\label{fig:youtube_lf_diversity}
\end{figure}

\subsubsection{Diversity Measures}
A key factor influencing labeling function performance is how they interact with other labeling functions. 
As in ensembling, we want labelers that provide complimentary information and have low correlated error rates, which improves ensemble efficiency and enables combining many weak classifiers to achieve stronger classification performance.  
To gain insight into the diversity of prompted labeling functions, we compute metrics informed by ensemble diversity measures \cite{kuncheva:ml2003}. 
Given a pair of labelers, $i$ and $j$, we construct a 2x2 contingency table of vote counts for pairs of unlabeled examples. 
In binary classification, where $N^{ij}$ is the total number of label pairs emitted by labelers $i$ and $j$, this table contains $N^{00} + N^{10} + N^{01} + N^{11}$ covered instances.
We consider the following diversity measures defined using these counts, normalizing all measures by the total size of the unlabeled training set.
\\ \\ \\ \\
\begin{enumerate}
    \item \emph{Agreement} $:= N^{00} + N^{11}$
    \item \emph{Disagreement} $:= N^{10} + N^{01}$
    \item \emph{Double Fault} $:= N^{00}$
    \item \emph{Double Correct} $:= N^{11}$
\end{enumerate}
Agreement and disagreement provide measures of correlation between two labeling function prompts and enable characterizing the degree to which prompts provide complimentary label information.

Figure \ref{fig:youtube_lf_diversity} shows a heatmap view of pairwise diversity of the YouTube dataset. 
Note there is more variation (disagreement) in the \tzeropp~models and less agreement (double fault and double correct) compared to the InstructGPT family of models. 
The Babbage model, for example, generates strongly correlated labels and less variation in label signal. 
\tzeropp~has higher variation in labels and less correlated errors across both classes. 
Lower correlated errors suggests that prompts evaluated using \tzeropp~are providing complimentary label information, resulting in greater ensemble efficiency and improving overall model performance \cite{gontijo-lopes:iclr2022}.
Similar patterns are observed in the other datasets (see Appendix \S \ref{sec:diversity}).

\section{Discussion and Conclusion}

Developing flexible methods to query and adapt large-scale foundation models for downstream tasks is emerging as a critical component of machine learning systems. 
Our work demonstrates several benefits of using prompted weak supervision to query and repurpose information found in language models. 
Combining multiple prompted labeling functions provides significant improvements over underspecified prompts commonly used for zero-shot classification.
By formulating tasks using multiple prompts, prompted weak supervision provides an inspectable mechanism for contextualizing task insight and querying knowledge found in large language models. 

Prompts provide several advantages that compliment traditional code-based labeling functions.
Unlike code, which is static and potentially expensive to refine, prompts are interpreted by an underlying language model, meaning the labels generated by prompts may improve as language models themselves continue improving.
Moreover, the prompts explored in this work likely underestimate the potential performance of our approach, as we focused on translating existing labeling functions rather than developing and refining new prompts.

In our experiments, \tzeropp, which was pretrained with multi-task prompted examples, consistently outperforms the InstructGPT family of language models when used for prompted weak supervision.
Future work may consider methods of generating additional prompted pretraining data that aligns more closely with how SMEs approach prompt design in weakly supervised workflows.
This is a particularly exciting use of data exhaust, as the process of querying and interacting with a language model can be used to directly improve the quality of the underlying model \cite{ouyang:corr22}. 

Finally, the success of contextual calibration underscores the benefits and current limitations of recalibration methods for prompt-based zero-shot learning. 
Performance gains, while consistent at the level of collections of prompts, is inconsistent and brittle at the level of an individual prompt. 
As new methods continue to improve language model calibration, we expect prompted weak supervision to benefit by increasing the ability of SMEs to refine the operating threshold of individual labeling functions.

\subsection*{Acknowledgements}
The authors would like to thank the rest of the research team at Snorkel AI for the many helpful conversations and feedback on this work.
Figures~\ref{fig:overview} and~\ref{fig:system} incorporate \href{https://commons.wikimedia.org/wiki/File:Scientist.svg}{this image} by Viktorvoight (\href{https://creativecommons.org/licenses/by-sa/3.0/deed.en}{CC BY-SA 3.0}).
Disclosure: Jason Fries and Stephen Bach contributed to this work as advisors to Snorkel AI.

\bibliographystyle{plainnat}
\bibliography{main}

\begin{thebibliography}{68}
\providecommand{\natexlab}[1]{#1}
\providecommand{\url}[1]{\texttt{#1}}
\expandafter\ifx\csname urlstyle\endcsname\relax
  \providecommand{\doi}[1]{doi: #1}\else
  \providecommand{\doi}{doi: \begingroup \urlstyle{rm}\Url}\fi

\bibitem[Arachie and Huang(2021{\natexlab{a}})]{arachie:jmlr21}
Chidubem Arachie and Bert Huang.
\newblock A general framework for adversarial label learning.
\newblock \emph{Journal of Machine Learning Research}, 22\penalty0
  (118):\penalty0 1--33, 2021{\natexlab{a}}.

\bibitem[Arachie and Huang(2021{\natexlab{b}})]{arachie:uai21}
Chidubem Arachie and Bert Huang.
\newblock Constrained labeling for weakly supervised learning.
\newblock In \emph{Uncertainty in Aritficial Intelligence (UAI)},
  2021{\natexlab{b}}.

\bibitem[Authors(2022)]{anonymous:arr22}
Anonymous Authors.
\newblock Prompt consistency for zero-shot task generalization.
\newblock Submitted to ACL Rolling Review, 2022.
\newblock URL \url{https://openreview.net/pdf?id=Ig8xeTpEmHf}.

\bibitem[Bach et~al.(2017)Bach, He, Ratner, and R\'e]{bach:icml17}
Stephen~H. Bach, Bryan He, Alexander Ratner, and Christopher R\'e.
\newblock Learning the structure of generative models without labeled data.
\newblock In \emph{International Conference on Machine Learning (ICML)}, 2017.

\bibitem[Bach et~al.(2019)Bach, Rodriguez, Liu, Luo, Shao, Xia, Sen, Ratner,
  Hancock, Alborzi, Kuchhal, R{\'e}, and Malkin]{bach:sigmod19-industrial}
Stephen~H. Bach, Daniel Rodriguez, Yintao Liu, Chong Luo, Haidong Shao,
  Cassandra Xia, Souvik Sen, Alexander Ratner, Braden Hancock, Houman Alborzi,
  Rahul Kuchhal, Christopher R{\'e}, and Rob Malkin.
\newblock Snorkel {DryBell}: {A} case study in deploying weak supervision at
  industrial scale.
\newblock In \emph{ACM SIGMOD Conference on Management of Data (SIGMOD)
  Industry Track}, 2019.

\bibitem[Biegel et~al.(2021)Biegel, El-Khatib, Oliveira, Baak, and
  Aben]{biegel:iclrws21}
Samantha Biegel, Rafah El-Khatib, Luiz Otavio Vilas~Boas Oliveira, Max Baak,
  and Nanne Aben.
\newblock Active weasul: Improving weak supervision with active learning.
\newblock In \emph{ICLR Workshop on Weakly Supervised Learning}, 2021.

\bibitem[Bommasani et~al.(2021)Bommasani, Hudson, Adeli, Altman, Arora, von
  Arx, Bernstein, Bohg, Bosselut, Brunskill, et~al.]{bommasani:arxiv21}
Rishi Bommasani, Drew~A Hudson, Ehsan Adeli, Russ Altman, Simran Arora, Sydney
  von Arx, Michael~S Bernstein, Jeannette Bohg, Antoine Bosselut, Emma
  Brunskill, et~al.
\newblock On the opportunities and risks of foundation models.
\newblock \emph{arXiv preprint arXiv:2108.07258}, 2021.

\bibitem[Bonifacio et~al.(2022)Bonifacio, Abonizio, Fadaee, and
  Nogueira]{bonifacio:arxiv22}
Luiz Bonifacio, Hugo Abonizio, Marzieh Fadaee, and Rodrigo Nogueira.
\newblock {InPars}: {D}ata augmentation for information retrieval using large
  language models.
\newblock \emph{arXiv preprint arXiv:2202.05144}, 2022.

\bibitem[Bringer et~al.(2019)Bringer, Israeli, Shoham, Ratner, and
  R{\'e}]{bringer:deem19}
Eran Bringer, Abraham Israeli, Yoav Shoham, Alex Ratner, and Christopher
  R{\'e}.
\newblock Osprey: Weak supervision of imbalanced extraction problems without
  code.
\newblock In \emph{International Workshop on Data Management for End-to-End
  Machine Learning (DEEM)}, 2019.

\bibitem[Brown et~al.(2020)Brown, Mann, Ryder, Subbiah, Kaplan, Dhariwal,
  Neelakantan, Shyam, Sastry, Askell, et~al.]{brown:neurips20}
Tom Brown, Benjamin Mann, Nick Ryder, Melanie Subbiah, Jared~D Kaplan, Prafulla
  Dhariwal, Arvind Neelakantan, Pranav Shyam, Girish Sastry, Amanda Askell,
  et~al.
\newblock Language models are few-shot learners.
\newblock \emph{Neural Information Processing Systems (NeurIPS)}, 2020.

\bibitem[Brust et~al.(2020)Brust, K{\"a}ding, and Denzler]{brust:ki20}
Clemens-Alexander Brust, Christoph K{\"a}ding, and Joachim Denzler.
\newblock Active and incremental learning with weak supervision.
\newblock \emph{KI-K{\"u}nstliche Intelligenz}, 34\penalty0 (2):\penalty0
  165--180, 2020.

\bibitem[Callahan et~al.(2019)Callahan, Fries, R{\'e}, Huddleston, Giori, Delp,
  and Shah]{callahan:npjdigmed19}
Alison Callahan, Jason~A Fries, Christopher R{\'e}, James~I Huddleston,
  Nicholas~J Giori, Scott Delp, and Nigam~H Shah.
\newblock Medical device surveillance with electronic health records.
\newblock \emph{NPJ digital medicine}, 2\penalty0 (1):\penalty0 1--10, 2019.

\bibitem[Chen et~al.(2022)Chen, Fu, Adila, Zhang, Sala, Fatahalian, and
  R{\'e}]{chen:arxiv22}
Mayee~F Chen, Daniel~Y Fu, Dyah Adila, Michael Zhang, Frederic Sala, Kayvon
  Fatahalian, and Christopher R{\'e}.
\newblock Shoring up the foundations: Fusing model embeddings and weak
  supervision.
\newblock \emph{arXiv preprint arXiv:2203.13270}, 2022.

\bibitem[Chia et~al.(2022)Chia, Bing, Poria, and Si]{chia:acl22}
Yew~Ken Chia, Lidong Bing, Soujanya Poria, and Luo Si.
\newblock {RelationPrompt}: {L}everaging prompts to generate synthetic data for
  zero-shot relation triplet extraction.
\newblock In \emph{Findings of the Association for Computational Linguistics},
  2022.

\bibitem[Craven et~al.(1999)Craven, Kumlien, et~al.]{craven:ismb99}
Mark Craven, Johan Kumlien, et~al.
\newblock Constructing biological knowledge bases by extracting information
  from text sources.
\newblock In \emph{Intelligent Systems for Molecular Biology (ISMB)}, 1999.

\bibitem[Dawid and Skene(1979)]{dawid:royalstats79}
A.~P. Dawid and A.~M. Skene.
\newblock Maximum likelihood estimation of observer error-rates using the {EM}
  algorithm.
\newblock \emph{Journal of the Royal Statistical Society C}, 28\penalty0
  (1):\penalty0 20--28, 1979.

\bibitem[Devlin et~al.(2019)Devlin, Chang, Lee, and Toutanova]{devlin:naacl19}
Jacob Devlin, Ming-Wei Chang, Kenton Lee, and Kristina Toutanova.
\newblock {BERT}: {P}re-training of deep bidirectional transformers for
  language understanding.
\newblock In \emph{Meeting of the North American Association for Computational
  Linguistics (NAACL)}, 2019.

\bibitem[Dunnmon et~al.(2020)Dunnmon, Ratner, Saab, Khandwala, Markert,
  Sagreiya, Goldman, Lee-Messer, Lungren, Rubin, et~al.]{dunnmon:patterns20}
Jared~A Dunnmon, Alexander~J Ratner, Khaled Saab, Nishith Khandwala, Matthew
  Markert, Hersh Sagreiya, Roger Goldman, Christopher Lee-Messer, Matthew~P
  Lungren, Daniel~L Rubin, et~al.
\newblock Cross-modal data programming enables rapid medical machine learning.
\newblock \emph{Patterns}, 1\penalty0 (2), 2020.

\bibitem[Elazar et~al.(2021)Elazar, Kassner, Ravfogel, Ravichander, Hovy,
  Sch{\"u}tze, and Goldberg]{elazar:tacl21}
Yanai Elazar, Nora Kassner, Shauli Ravfogel, Abhilasha Ravichander, Eduard
  Hovy, Hinrich Sch{\"u}tze, and Yoav Goldberg.
\newblock Measuring and improving consistency in pretrained language models.
\newblock \emph{Transactions of the Association for Computational Linguistics},
  9:\penalty0 1012--1031, 2021.

\bibitem[Eyuboglu et~al.(2021)Eyuboglu, Angus, Patel, Pareek, Davidzon, Long,
  Dunnmon, and Lungren]{eyuboglu:naturecommunications21}
Sabri Eyuboglu, Geoffrey Angus, Bhavik~N Patel, Anuj Pareek, Guido Davidzon,
  Jin Long, Jared Dunnmon, and Matthew~P Lungren.
\newblock Multi-task weak supervision enables anatomically-resolved abnormality
  detection in whole-body fdg-pet/ct.
\newblock \emph{Nature communications}, 12\penalty0 (1):\penalty0 1--15, 2021.

\bibitem[Fries et~al.(2021)Fries, Steinberg, Khattar, Fleming, Posada,
  Callahan, and Shah]{fries:nc2021}
Jason~A Fries, Ethan Steinberg, Saelig Khattar, Scott~L Fleming, Jose Posada,
  Alison Callahan, and Nigam~H Shah.
\newblock Ontology-driven weak supervision for clinical entity classification
  in electronic health records.
\newblock \emph{Nature communications}, 12\penalty0 (1):\penalty0 1--11, 2021.

\bibitem[Fu et~al.(2020)Fu, Chen, Sala, Hooper, Fatahalian, and
  R{\'e}]{fu:icml2020}
Daniel Fu, Mayee Chen, Frederic Sala, Sarah Hooper, Kayvon Fatahalian, and
  Christopher R{\'e}.
\newblock Fast and three-rious: Speeding up weak supervision with triplet
  methods.
\newblock In \emph{International Conference on Machine Learning}, pages
  3280--3291. PMLR, 2020.

\bibitem[Gao et~al.(2020)Gao, Biderman, Black, Golding, Hoppe, Foster, Phang,
  He, Thite, Nabeshima, et~al.]{gao:arxiv20}
Leo Gao, Stella Biderman, Sid Black, Laurence Golding, Travis Hoppe, Charles
  Foster, Jason Phang, Horace He, Anish Thite, Noa Nabeshima, et~al.
\newblock The pile: {A}n 800{GB} dataset of diverse text for language modeling.
\newblock \emph{arXiv preprint arXiv:2101.00027}, 2020.

\bibitem[Gontijo-Lopes et~al.(2022)Gontijo-Lopes, Dauphin, and
  Cubuk]{gontijo-lopes:iclr2022}
Raphael Gontijo-Lopes, Yann Dauphin, and Ekin~Dogus Cubuk.
\newblock No one representation to rule them all: Overlapping features of
  training methods.
\newblock In \emph{International Conference on Learning Representations}, 2022.
\newblock URL \url{https://openreview.net/forum?id=BK-4qbGgIE3}.

\bibitem[Guo et~al.(2017)Guo, Pleiss, Sun, and Weinberger]{guo:icml17}
Chuan Guo, Geoff Pleiss, Yu~Sun, and Kilian~Q. Weinberger.
\newblock On calibration of modern neural networks.
\newblock In \emph{{ICML}}, volume~70 of \emph{Proceedings of Machine Learning
  Research}, pages 1321--1330. {PMLR}, 2017.

\bibitem[Jiang et~al.(2020)Jiang, Xu, Araki, and Neubig]{jiang:tacl20}
Zhengbao Jiang, Frank~F. Xu, Jun Araki, and Graham Neubig.
\newblock How can we know what language models know?
\newblock \emph{Transactions of the Association for Computational Linguistics},
  8:\penalty0 423--438, 2020.
\newblock \doi{10.1162/tacl_a_00324}.
\newblock URL \url{https://aclanthology.org/2020.tacl-1.28}.

\bibitem[Karamanolakis et~al.(2021)Karamanolakis, Mukherjee, Zheng, and
  Awadallah]{karamanolakis:naacl21}
Giannis Karamanolakis, Subhabrata Mukherjee, Guoqing Zheng, and Ahmed~Hassan
  Awadallah.
\newblock Self-training with weak supervision.
\newblock In \emph{Meeting of the North American Association for Computational
  Linguistics (NAACL)}, 2021.

\bibitem[Kuang et~al.(2021)Kuang, Arachie, Liang, Narayana, DeSalvo, Quinn,
  Huang, Downs, and Yang]{kuang:aistats22}
Zhaobin Kuang, Chidubem Arachie, Bangyong Liang, Pradyumna Narayana, Giulia
  DeSalvo, Michael Quinn, Bert Huang, Geoffrey Downs, and Yang Yang.
\newblock Firebolt: {W}eak supervision under weaker assumptions.
\newblock In \emph{Artificial Intelligence and Statistics (AISTATS)}, 2021.

\bibitem[Kuncheva and Whitaker(2003)]{kuncheva:ml2003}
Ludmila~I Kuncheva and Christopher~J Whitaker.
\newblock Measures of diversity in classifier ensembles and their relationship
  with the ensemble accuracy.
\newblock \emph{Machine learning}, 51\penalty0 (2):\penalty0 181--207, 2003.

\bibitem[Lang et~al.(2022)Lang, Agrawal, Kim, and Sontag]{lang:arxiv22}
Hunter Lang, Monica Agrawal, Yoon Kim, and David Sontag.
\newblock Co-training improves prompt-based learning for large language models.
\newblock \emph{arXiv preprint arXiv:2202.00828}, 2022.

\bibitem[Liu et~al.(2021)Liu, Yuan, Fu, Jiang, Hayashi, and
  Neubig]{liu:arxiv21}
Pengfei Liu, Weizhe Yuan, Jinlan Fu, Zhengbao Jiang, Hiroaki Hayashi, and
  Graham Neubig.
\newblock Pre-train, prompt, and predict: A systematic survey of prompting
  methods in natural language processing.
\newblock \emph{arXiv preprint arXiv:2107.13586}, 2021.

\bibitem[Liu et~al.(2020)Liu, Ott, Goyal, Du, Joshi, Chen, Levy, Lewis,
  Zettlemoyer, and Stoyanov]{liu:openreview2020}
Yinhan Liu, Myle Ott, Naman Goyal, Jingfei Du, Mandar Joshi, Danqi Chen, Omer
  Levy, Mike Lewis, Luke Zettlemoyer, and Veselin Stoyanov.
\newblock Ro{\{}bert{\}}a: A robustly optimized {\{}bert{\}} pretraining
  approach, 2020.
\newblock URL \url{https://openreview.net/forum?id=SyxS0T4tvS}.

\bibitem[Mazzetto et~al.(2021{\natexlab{a}})Mazzetto, Cousins, Sam, Bach, and
  Upfal]{mazzetto:icml21}
A.~Mazzetto, C.~Cousins, D.~Sam, S.~H. Bach, and E.~Upfal.
\newblock Adversarial multiclass learning under weak supervision with
  performance guarantees.
\newblock In \emph{International Conference on Machine Learning (ICML)},
  2021{\natexlab{a}}.

\bibitem[Mazzetto et~al.(2021{\natexlab{b}})Mazzetto, Sam, Park, Upfal, and
  Bach]{mazzetto:aistats21}
A.~Mazzetto, D.~Sam, A.~Park, E.~Upfal, and S.~H. Bach.
\newblock Semi-supervised aggregation of dependent weak supervision sources
  with performance guarantees.
\newblock In \emph{Artificial Intelligence and Statistics (AISTATS)},
  2021{\natexlab{b}}.

\bibitem[Mintz et~al.(2009)Mintz, Bills, Snow, and Jurafsky]{mintz:acl09}
M.~Mintz, S.~Bills, R.~Snow, and D.~Jurafsky.
\newblock Distant supervision for relation extraction without labeled data.
\newblock In \emph{Meeting of the Association for Computational Linguistics
  (ACL)}, 2009.

\bibitem[Mishra et~al.(2022)Mishra, Khashabi, Baral, and
  Hajishirzi]{mishra:acl22}
Swaroop Mishra, Daniel Khashabi, Chitta Baral, and Hannaneh Hajishirzi.
\newblock Cross-task generalization via natural language crowdsourcing
  instruction.
\newblock In \emph{Meeting of the Association for Computational Linguistics
  (ACL)}, 2022.

\bibitem[Ouyang et~al.(2022)Ouyang, Wu, Jiang, Almeida, Wainwright, Mishkin,
  Zhang, Agarwal, Slama, Ray, Schulman, Hilton, Kelton, Miller, Simens, Askell,
  Welinder, Christiano, Leike, and Lowe]{ouyang:corr22}
Long Ouyang, Jeff Wu, Xu~Jiang, Diogo Almeida, Carroll~L. Wainwright, Pamela
  Mishkin, Chong Zhang, Sandhini Agarwal, Katarina Slama, Alex Ray, John
  Schulman, Jacob Hilton, Fraser Kelton, Luke Miller, Maddie Simens, Amanda
  Askell, Peter Welinder, Paul~F. Christiano, Jan Leike, and Ryan Lowe.
\newblock Training language models to follow instructions with human feedback.
\newblock \emph{CoRR}, abs/2203.02155, 2022.

\bibitem[Platt et~al.(1999)]{platt:book99}
John Platt et~al.
\newblock Probabilistic outputs for support vector machines and comparisons to
  regularized likelihood methods.
\newblock \emph{Advances in large margin classifiers}, 10\penalty0
  (3):\penalty0 61--74, 1999.

\bibitem[Rae et~al.(2021)Rae, Borgeaud, Cai, Millican, Hoffmann, Song,
  Aslanides, Henderson, Ring, Young, et~al.]{rae:arxiv21}
Jack~W Rae, Sebastian Borgeaud, Trevor Cai, Katie Millican, Jordan Hoffmann,
  Francis Song, John Aslanides, Sarah Henderson, Roman Ring, Susannah Young,
  et~al.
\newblock Scaling language models: {M}ethods, analysis \& insights from
  training gopher.
\newblock \emph{arXiv preprint arXiv:2112.11446}, 2021.

\bibitem[Raffel et~al.(2020)Raffel, Shazeer, Roberts, Lee, Narang, Matena,
  Zhou, Li, and Liu]{raffel:jmlr20}
Colin Raffel, Noam Shazeer, Adam Roberts, Katherine Lee, Sharan Narang, Michael
  Matena, Yanqi Zhou, Wei Li, and Peter~J. Liu.
\newblock Exploring the limits of transfer learning with a unified text-to-text
  transformer.
\newblock \emph{Journal of Machine Learning Research}, 21\penalty0
  (140):\penalty0 1--67, 2020.

\bibitem[Ratner et~al.(2020)Ratner, Bach, Ehrenberg, Fries, Wu, and
  R{\'e}]{ratner:vldbj20}
A.~J. Ratner, S.~H. Bach, H.~E. Ehrenberg, J.~Fries, S.~Wu, and C.~R{\'e}.
\newblock Snorkel: {R}apid training data creation with weak supervision.
\newblock \emph{The VLDB Journal}, 29\penalty0 (2):\penalty0 709--730, 2020.

\bibitem[Ratner et~al.(2016)Ratner, De~Sa, Wu, Selsam, and
  R{\'e}]{ratner:neurips16}
Alexander~J Ratner, Christopher~M De~Sa, Sen Wu, Daniel Selsam, and Christopher
  R{\'e}.
\newblock Data programming: {C}reating large training sets, quickly.
\newblock In \emph{Neural Information Processing Systems (NeurIPS)}, 2016.

\bibitem[Ratner et~al.(2019)Ratner, Hancock, Dunnmon, Sala, Pandey, and
  R{\'e}]{ratner:aaai19}
Alexander~J Ratner, Braden Hancock, Jared Dunnmon, Frederic Sala, Shreyash
  Pandey, and Christopher R{\'e}.
\newblock Training complex models with multi-task weak supervision.
\newblock In \emph{AAAI Conference on Artificial Intelligence (AAAI)}, 2019.

\bibitem[Safranchik et~al.(2020)Safranchik, Luo, and Bach]{safranchik:aaai20}
Esteban Safranchik, Shiying Luo, and Stephen~H. Bach.
\newblock Weakly supervised sequence tagging from noisy rules.
\newblock In \emph{AAAI Conference on Artificial Intelligence (AAAI)}, 2020.

\bibitem[Sala et~al.(2019)Sala, Varma, Sagawa, Fries, Fu, Khattar, Ramamoorthy,
  Xiao, Fatahalian, Priest, et~al.]{sala:neurips19}
Frederic Sala, Paroma Varma, Shiori Sagawa, Jason Fries, Daniel Fu, Saelig
  Khattar, Ashwini Ramamoorthy, Ke~Xiao, Kayvon Fatahalian, James Priest,
  et~al.
\newblock Multi-resolution weak supervision for sequential data.
\newblock In \emph{Neural Information Processing Systems (NeurIPS}, 2019.

\bibitem[Sanh et~al.(2022)Sanh, Webson, Raffel, Bach, Sutawika, Alyafeai,
  Chaffin, Stiegler, Scao, Raja, Dey, Bari, Xu, Thakker, Sharma, Szczechla,
  Kim, Chhablani, Nayak, Datta, Chang, Jiang, Wang, Manica, Shen, Yong, Pandey,
  Bawden, Wang, Neeraj, Rozen, Sharma, Santilli, Fevry, Fries, Teehan,
  Biderman, Gao, Bers, Wolf, and Rush]{sanh:iclr22}
Victor Sanh, Albert Webson, Colin Raffel, Stephen~H. Bach, Lintang Sutawika,
  Zaid Alyafeai, Antoine Chaffin, Arnaud Stiegler, Teven~Le Scao, Arun Raja,
  Manan Dey, M~Saiful Bari, Canwen Xu, Urmish Thakker, Shanya Sharma, Eliza
  Szczechla, Taewoon Kim, Gunjan Chhablani, Nihal Nayak, Debajyoti Datta,
  Jonathan Chang, Mike Tian-Jian Jiang, Han Wang, Matteo Manica, Sheng Shen,
  Zheng~Xin Yong, Harshit Pandey, Rachel Bawden, Thomas Wang, Trishala Neeraj,
  Jos Rozen, Abheesht Sharma, Andrea Santilli, Thibault Fevry, Jason~Alan
  Fries, Ryan Teehan, Stella Biderman, Leo Gao, Tali Bers, Thomas Wolf, and
  Alexander~M. Rush.
\newblock Multitask prompted training enables zero-shot task generalization.
\newblock In \emph{International Conference on Learning Representations
  (ICLR)}, 2022.

\bibitem[Schick and Sch{\"u}tze(2021)]{schick:emnlp21}
Timo Schick and Hinrich Sch{\"u}tze.
\newblock Generating datasets with pretrained language models.
\newblock In \emph{Conference on Empirical Methods in Natural Language
  Processing (EMNLP)}, 2021.

\bibitem[Shin et~al.(2022)Shin, Li, Vishwakarma, Roberts, and
  Sala]{shin:iclr22}
Changho Shin, Winfred Li, Harit Vishwakarma, Nicholas Roberts, and Frederic
  Sala.
\newblock Universalizing weak supervision.
\newblock In \emph{International Conference on Learning Representations
  (ICLR)}, 2022.

\bibitem[Shin et~al.(2020)Shin, Razeghi, Logan~IV, Wallace, and
  Singh]{shin:emnlp20}
Taylor Shin, Yasaman Razeghi, Robert~L Logan~IV, Eric Wallace, and Sameer
  Singh.
\newblock Auto{P}rompt: {E}liciting knowledge from language models with
  automatically generated prompts.
\newblock In \emph{Conference on Empirical Methods in Natural Language
  Processing (EMNLP)}, 2020.

\bibitem[Suri et~al.(2020)Suri, Chanda, Bulut, Narayana, Zeng, Bailis, Basu,
  Narlikar, R\'{e}, and Sethi]{suri:vldb20}
Sahaana Suri, Raghuveer Chanda, Neslihan Bulut, Pradyumna Narayana, Yemao Zeng,
  Peter Bailis, Sugato Basu, Girija Narlikar, Christopher R\'{e}, and Abishek
  Sethi.
\newblock Leveraging organizational resources to adapt models to new data
  modalities.
\newblock \emph{Proc. VLDB Endow.}, 13\penalty0 (12):\penalty0 3396–3410,
  2020.

\bibitem[Varma and R{\'e}(2018)]{varma:vldb18}
Paroma Varma and Christopher R{\'e}.
\newblock Snuba: {A}utomating weak supervision to label training data.
\newblock \emph{Proceedings of the VLDB Endowment}, 12\penalty0 (3):\penalty0
  223, 2018.

\bibitem[Varma et~al.(2016)Varma, He, Iter, Xu, Yu, De~Sa, and
  R{\'e}]{varma:arxiv16}
Paroma Varma, Bryan He, Dan Iter, Peng Xu, Rose Yu, Christopher De~Sa, and
  Christopher R{\'e}.
\newblock Socratic learning: Augmenting generative models to incorporate latent
  subsets in training data.
\newblock \emph{arXiv preprint arXiv:1610.08123}, 2016.

\bibitem[Varma et~al.(2019)Varma, Sala, He, Ratner, and R\'e]{varma:icml19}
Paroma Varma, Fred Sala, Ann He, Alex Ratner, and Christopher R\'e.
\newblock Learning dependency structures for weak supervision models.
\newblock In \emph{International Conference on Machine Learning (ICML)}, 2019.

\bibitem[Vaswani et~al.(2017)Vaswani, Shazeer, Parmar, Uszkoreit, Jones, Gomez,
  Kaiser, and Polosukhin]{vaswani:neurips17}
Ashish Vaswani, Noam Shazeer, Niki Parmar, Jakob Uszkoreit, Llion Jones,
  Aidan~N Gomez, {\L}ukasz Kaiser, and Illia Polosukhin.
\newblock Attention is all you need.
\newblock In \emph{Neural Information Processing Systems (NeurIPS)}, 2017.

\bibitem[Wang et~al.(2022)Wang, Wei, Schuurmans, Le, Chi, and
  Zhou]{wang:arxiv22}
Xuezhi Wang, Jason Wei, Dale Schuurmans, Quoc Le, Ed~Chi, and Denny Zhou.
\newblock Self-consistency improves chain of thought reasoning in language
  models.
\newblock \emph{arXiv preprint arXiv:2203.11171}, 2022.

\bibitem[Webson and Pavlick(2021)]{webson:arxiv21}
Albert Webson and Ellie Pavlick.
\newblock Do prompt-based models really understand the meaning of their
  prompts?
\newblock \emph{arXiv preprint arXiv:2109.01247}, 2021.

\bibitem[Wei et~al.(2022{\natexlab{a}})Wei, Bosma, Zhao, Guu, Yu, Lester, Du,
  Dai, and Le]{wei:iclr22}
Jason Wei, Maarten Bosma, Vincent~Y Zhao, Kelvin Guu, Adams~Wei Yu, Brian
  Lester, Nan Du, Andrew~M Dai, and Quoc~V Le.
\newblock Finetuned language models are zero-shot learners.
\newblock In \emph{International Conference on Learning Representations
  (ICLR)}, 2022{\natexlab{a}}.

\bibitem[Wei et~al.(2022{\natexlab{b}})Wei, Wang, Schuurmans, Bosma, Chi, Le,
  and Zhou]{wei:arxiv22}
Jason Wei, Xuezhi Wang, Dale Schuurmans, Maarten Bosma, Ed~Chi, Quoc Le, and
  Denny Zhou.
\newblock Chain of thought prompting elicits reasoning in large language
  models.
\newblock \emph{arXiv preprint arXiv:2201.11903}, 2022{\natexlab{b}}.

\bibitem[Wu et~al.(2022)Wu, Gardner, Stenetorp, and Dasigi]{wu:acl22}
Yuxiang Wu, Matt Gardner, Pontus Stenetorp, and Pradeep Dasigi.
\newblock Generating data to mitigate spurious correlations in natural language
  inference datasets.
\newblock In \emph{Meeting of the Association for Computational Linguistics
  (ACL)}, 2022.

\bibitem[Ye et~al.(2022)Ye, Gao, Li, Xu, Feng, Wu, Yu, and Kong]{ye:arxiv22}
Jiacheng Ye, Jiahui Gao, Qintong Li, Hang Xu, Jiangtao Feng, Zhiyong Wu, Tao
  Yu, and Lingpeng Kong.
\newblock {ZeroGen}: {E}fficient zero-shot learning via dataset generation.
\newblock \emph{arXiv preprint arXiv:2202.07922}, 2022.

\bibitem[Yu et~al.(2022)Yu, Ding, and Bach]{yu:aistats22}
P.~Yu, T.~Ding, and S.~H. Bach.
\newblock Learning from multiple noisy partial labelers.
\newblock In \emph{Artificial Intelligence and Statistics (AISTATS)}, 2022.

\bibitem[Yu et~al.(2021)Yu, Zuo, Jiang, Ren, Zhao, and Zhang]{yu:naacl21}
Yue Yu, Simiao Zuo, Haoming Jiang, Wendi Ren, Tuo Zhao, and Chao Zhang.
\newblock Fine-tuning pre-trained language model with weak supervision: A
  contrastive-regularized self-training approach.
\newblock In \emph{Conference of the North American Chapter of the Association
  for Computational Linguistics (NAACL)}, 2021.

\bibitem[Zelikman et~al.(2022)Zelikman, Wu, and Goodman]{zelikman:arxiv22}
Eric Zelikman, Yuhuai Wu, and Noah~D Goodman.
\newblock {STaR}: {B}ootstrapping reasoning with reasoning.
\newblock \emph{arXiv preprint arXiv:2203.14465}, 2022.

\bibitem[Zhang et~al.(2021)Zhang, Yu, Li, Wang, Yang, Yang, and
  Ratner]{zhang:neurips21}
Jieyu Zhang, Yue Yu, Yinghao Li, Yujing Wang, Yaming Yang, Mao Yang, and
  Alexander Ratner.
\newblock {WRENCH}: A comprehensive benchmark for weak supervision.
\newblock In \emph{Thirty-fifth Conference on Neural Information Processing
  Systems Datasets and Benchmarks Track (Round 2)}, 2021.
\newblock URL \url{https://openreview.net/forum?id=Q9SKS5k8io}.

\bibitem[Zhang et~al.(2022{\natexlab{a}})Zhang, Hsieh, Yu, Zhang, and
  Ratner]{zhang:arxiv22}
Jieyu Zhang, Cheng-Yu Hsieh, Yue Yu, Chao Zhang, and Alexander Ratner.
\newblock A survey on programmatic weak supervision.
\newblock \emph{arXiv preprint arXiv:2202.05433}, 2022{\natexlab{a}}.

\bibitem[Zhang et~al.(2022{\natexlab{b}})Zhang, Wang, Song, Wang, Yang, Bai,
  and Ratner]{zhang:iclr22}
Jieyu Zhang, Bohan Wang, Xiangchen Song, Yujing Wang, Yaming Yang, Jing Bai,
  and Alexander Ratner.
\newblock Creating training sets via weak indirect supervision.
\newblock In \emph{International Conference on Learning Representations
  (ICLR)}, 2022{\natexlab{b}}.

\bibitem[Zhang et~al.(2022{\natexlab{c}})Zhang, Yu, Shetty, Song, and
  Zhang]{zhang:acl22}
Rongzhi Zhang, Yue Yu, Pranav Shetty, Le~Song, and Chao Zhang.
\newblock {PRBoost}: {P}rompt-based rule discovery and boosting for interactive
  weakly-supervised learning.
\newblock In \emph{Meeting of the Association for Computational Linguistics
  (ACL)}, 2022{\natexlab{c}}.

\bibitem[Zhao et~al.(2021)Zhao, Wallace, Feng, Klein, and Singh]{zhao:icml21}
Zihao Zhao, Eric Wallace, Shi Feng, Dan Klein, and Sameer Singh.
\newblock Calibrate before use: Improving few-shot performance of language
  models.
\newblock In \emph{{ICML}}, volume 139 of \emph{Proceedings of Machine Learning
  Research}, pages 12697--12706. {PMLR}, 2021.

\end{thebibliography}

\appendix

\pagebreak
\section{Appendix}

\subsection{GPT-3 API Costs}
\label{sec:cost}

%
%
\begin{table}[!ht]
\centering
\begin{tabular}{llrrrrr}
\toprule
\multicolumn{3}{c}{ } & \multicolumn{4}{c}{InstructGPT Language Models} \\  \cmidrule(lr){4-7} 
Dataset & Supervision & \#Queries & Ada     & Babbage & Curie    & DaVinci   \\
\midrule
YouTube & Zero Shot   & 1,586     & \$0.04  & \$0.06  & \$0.28   & \$2.76         \\
YouTube & Prompted WS & 10,586     & \$0.43  & \$0.65  & \$3.24   & \$32.40       \\
\midrule
SMS     & Zero Shot   & 4,571     & \$0.11  & \$0.16  & \$0.82   & \$8.24        \\
SMS     & Prompted WS & 333,683     & \$9.72  & \$14.59 & \$72.93  & \$729.31      \\
\midrule
Spouse  & Zero Shot   & 22,254    & \$1.52  & \$2.28  & \$11.40  & \$113.97     \\
Spouse  & Prompted WS & 200,286    & \$16.02 & \$24.03 & \$120.16 & \$1,201.62   \\
\bottomrule
\end{tabular}
\caption{OpenAI API estimated query costs for labeling WRENCH training sets with InstructGPT family of language models. See https://openai.com/api/pricing/ (accessed 03/01/2022).}
\label{tab:api_costs}
\end{table}

\subsection{Zero Shot Prompt Baseline}

\subsubsection{End Model Generalization} Table \ref{tab:baselines_v1} contains performance of Zero Shot (ZS) prompts directly evaluated on test data compared to the same prompts used for prompted weak supervision, where we programmatically label the training split, train a RoBERTa end model, and evaluate on test data (ZS+End Model).
All prompts are contextually calibrated. 
The RoBERTa end model provides consistent improvements. 

\begin{table}[!ht]
\centering
\begin{tabular}{ l cc cc cc }
\toprule
& \multicolumn{2}{c}{Youtube (Accuracy)} & \multicolumn{2}{c}{SMS (F1)} & \multicolumn{2}{c}{Spouse (F1)} \\ \cmidrule(lr){2-3}\cmidrule(lr){4-5}\cmidrule(lr){6-7}
        & ZS         & ZS+End Model           & ZS        & ZS+End Model           & ZS        & ZS+End Model           \\ \midrule
\tzeropp    & 55.6 (0.0)          & \textbf{58.7 (2.4)}          & 34.0 (0.0)         & \textbf{83.2 (2.4)}          & \textbf{63.0 (0.0)}          & 41.5 (13.1)         \\ 
Curie   & \textbf{54.4 (0.0) }         & 52.8 (0.0)          &  0.0 (0.0)         & 0.0 (0.0)           & 38.3 (0.0)          & \textbf{49.6 (1.0)}            \\ 
Babbage & 55.6 (0.0)          & \textbf{78.5 (3.0) }         & 20.6 (0.0)         & \textbf{32.2 (3.0)}          & 26.9 (0.0)          & \textbf{40.9 (0.9)} \\ 
Ada     & 44.8 (0.0)          & \textbf{51.7 (2.4) }         & 25.1 (0.0)         & \textbf{26.3 (2.6) }         & 17.2 (0.0)          & \textbf{19.1 (0.8)}  \\ 
\bottomrule
\end{tabular}
\caption{Comparing the Zero Shot (ZS) prompt as a direct classification model for test data versus the same prompt when used as a labeler to programmatically generate training data for a RoBERTa model (ZS+End Model). The best performing prompt performances are in bold.}
\label{tab:baselines_v1}
\end{table}

\begin{table}[!ht]
\centering
\begin{tabular}{lllll|ll}
\toprule
Dataset & Prompts &      Language Model &   Precision &      Recall &         F1 &   Accuracy \\
\midrule
YouTube &  PWS+ZS &                \tzeropp &  92.3 (0.5) &  90.8 (0.9) & 91.0 (0.8) & 91.2 (0.8) \\
YouTube &     PWS &                \tzeropp &  92.6 (0.5) &  91.7 (0.5) & \textbf{91.9 (0.5)} & \textbf{92.0 (0.5)} \\ \midrule
    SMS &  PWS+ZS &                \tzeropp &  96.5 (0.8) &  92.8 (1.4) & \textbf{94.5 (0.4)} &\textbf{ 98.6 (0.1)} \\
    SMS &     PWS &                \tzeropp &  95.9 (2.5) &  88.1 (1.1) & 91.8 (1.6) & 97.9 (0.4) \\ \midrule
 Spouse &  PWS+ZS &                \tzeropp &  52.6 (1.0) &  75.4 (2.4) & 61.8 (0.8) & 92.5 (0.2) \\
 Spouse &     PWS &                \tzeropp &  54.2 (1.8) &  75.4 (1.2) & \textbf{62.9 (0.8)} & \textbf{92.8 (0.3)} \\ \midrule
YouTube &  PWS+ZS &   InstructGPT Curie &  80.5 (1.1) &  70.5 (1.1) & 68.9 (1.4) & 72.0 (1.0) \\
YouTube &     PWS &   InstructGPT Curie &  80.1 (1.0) &  77.1 (2.1) & \textbf{76.7 (2.3)} &\textbf{ 77.7 (1.9)} \\ \midrule
    SMS &  PWS+ZS &   InstructGPT Curie &  53.4 (5.8) &  80.6 (2.9) & 63.0 (4.9) & 86.0 (3.7) \\
    SMS &     PWS &   InstructGPT Curie & 60.6 (11.5) &  83.8 (4.4) & \textbf{65.7 (5.8)} & \textbf{86.2 (3.9)} \\ \midrule
 Spouse &  PWS+ZS &   InstructGPT Curie &  35.6 (2.2) &  58.9 (5.6) & \textbf{43.2 (0.8)} & \textbf{87.5 (1.2)} \\
 Spouse &     PWS &   InstructGPT Curie &  29.5 (0.9) &  67.5 (2.0) & 41.0 (0.9) & 84.3 (0.7) \\ \midrule
YouTube &  PWS+ZS & InstructGPT Babbage &  83.7 (0.6) &  83.0 (0.7) & 83.0 (0.6) & 83.1 (0.6) \\
YouTube &     PWS & InstructGPT Babbage &  85.8 (1.2) &  84.8 (1.4) & \textbf{84.9 (1.4)} & \textbf{85.1 (1.3)} \\ \midrule
    SMS &  PWS+ZS & InstructGPT Babbage &  13.4 (0.0) & 100.0 (0.0) & 23.6 (0.0) & 13.4 (0.0) \\
    SMS &     PWS & InstructGPT Babbage &  13.4 (0.0) & 100.0 (0.0) & 23.6 (0.0) & 13.4 (0.0) \\ \midrule
 Spouse &  PWS+ZS & InstructGPT Babbage &  25.0 (2.7) &  59.5 (4.0) & 34.3 (2.2) & \textbf{80.9 (2.4)} \\
 Spouse &     PWS & InstructGPT Babbage &  24.2 (2.2) &  67.7 (5.6) & \textbf{34.9 (1.7) }& 79.3 (1.9) \\ \midrule
YouTube &  PWS+ZS &     InstructGPT Ada &  54.0 (9.5) &  50.6 (0.3) & \textbf{36.0 (0.7)} & \textbf{53.3 (0.3)} \\
YouTube &     PWS &     InstructGPT Ada &  34.8 (8.4) &  50.1 (0.1) & 34.7 (0.2) & 52.9 (0.1) \\ \midrule
    SMS &  PWS+ZS &     InstructGPT Ada &  13.4 (0.0) & 100.0 (0.0) & 23.6 (0.0) & 13.4 (0.0) \\
    SMS &     PWS &     InstructGPT Ada &  16.6 (1.3) &  99.5 (0.5) & \textbf{28.3 (1.8)} & \textbf{30.4 (6.2) }\\ \midrule
 Spouse &  PWS+ZS &     InstructGPT Ada &  20.0 (0.4) &  53.1 (1.7) & \textbf{29.0 (0.3)} & 79.0 (0.8) \\
 Spouse &     PWS &     InstructGPT Ada &  16.8 (5.6) &  20.6 (8.1) & 17.7 (6.2) & \textbf{88.7 (1.4)} \\
\bottomrule
\end{tabular}
\caption{Incorporating the Zero Shot prompt as an additional labeling function in Prompted Weak Superision.}
\label{tab:prompt_results_pws_zs}
\end{table}

\subsubsection{Zero Shot Labeling Function} Table \ref{tab:prompt_results_pws_zs} contains results for prompted weak supervision models that add the Zero Shot prompt as an additional labeling function.
Performance benefits were mixed, with models generally negatively impacted by incorporating the Zero Shot labeler. 
Here \tzeropp~had an average improvement of 0.2 F1 points, while InstructGPT Curie and Babbage has an average drop of 2.8 and 0.8 F1 points respectively.
InstructGPT Ada improved by 2.6 F1 points on average. 

\subsection{Prompt Calibration}
\label{app:calibration}

\subsubsection{Contextual Calibration} 
We find calibration improves performance of prompted labeling functions, with the largest gains found in settings where uncalibrated prompts display pathological performance.
We observed that the InstructGPT family of language models performed very poorly in many zero shot and prompted weak supervision experiments, as shown in Table \ref{tab:prompt_results_uncalibrated}.
The performance benefits of contextual calibration for all language models and datasets are outlined for the Zero Shot baseline in Table \ref{tab:zs_cc_all} and for prompted weak supervision in Table \ref{tab:pws_cc_all}.

\begin{figure}[!ht]
    \centering
    \includegraphics[width=1.0\textwidth]{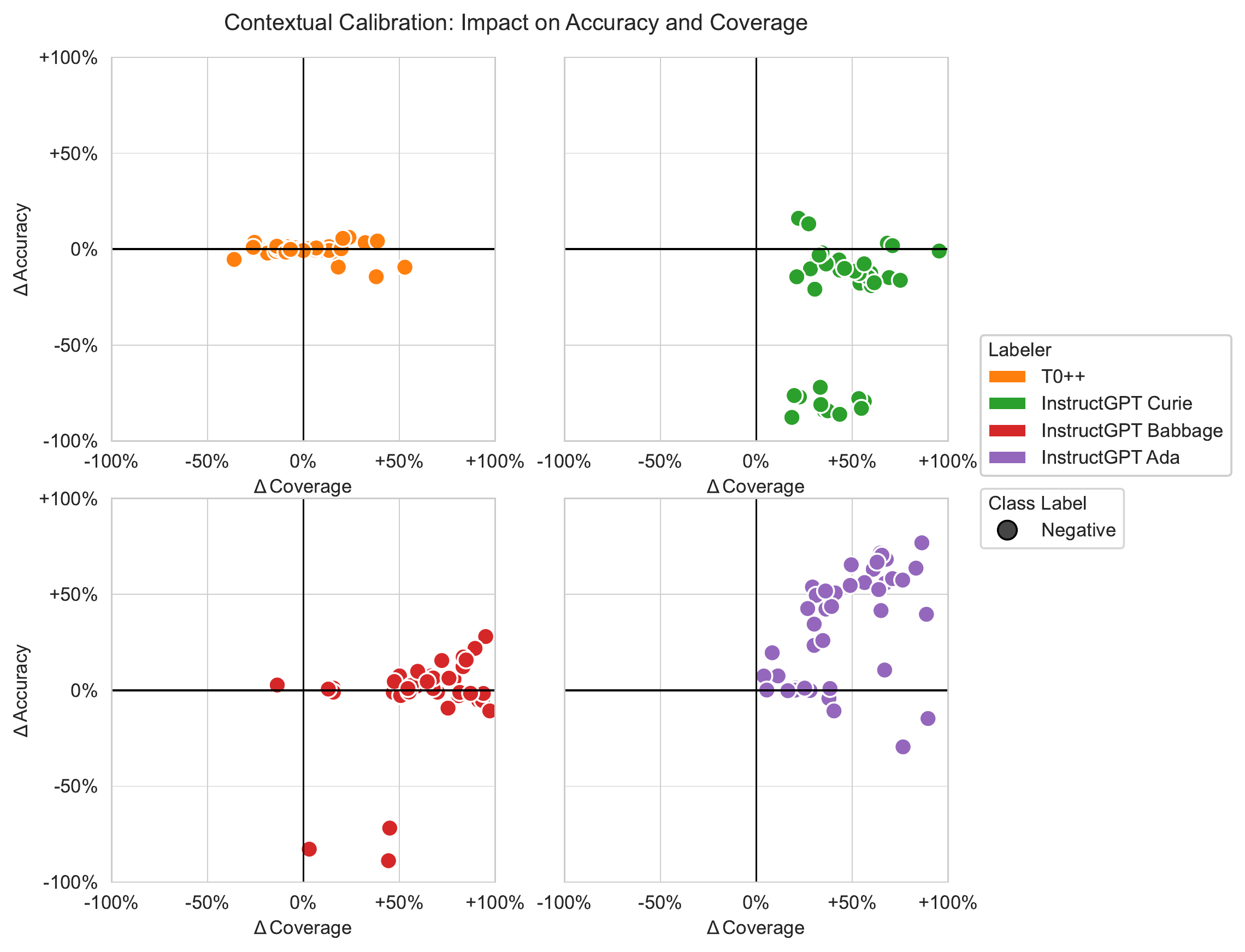}
    \caption{Accuracy and coverage changes as a result of contextual calibration, broken down by the negative class label.}
    \label{fig:0_class_calibration_deltas}
\end{figure}

\begin{figure}[!ht]
    \centering
    \includegraphics[width=1.0\textwidth]{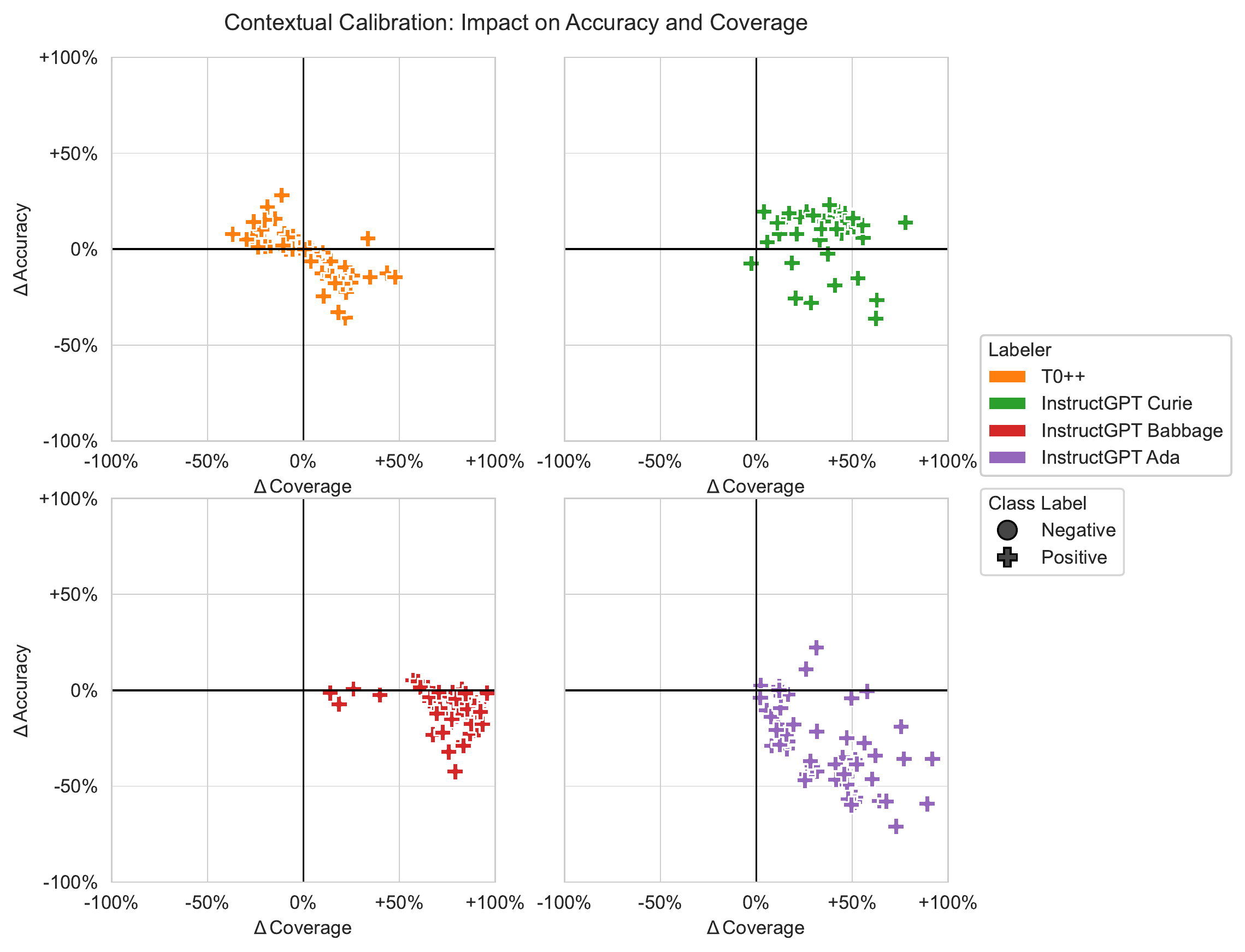}
    \caption{Accuracy and coverage changes as a result of contextual calibration, broken down by the positive class label.}
    \label{fig:1_class_calibration_deltas}
\end{figure}

Figures \ref{fig:0_class_calibration_deltas} and \ref{fig:1_class_calibration_deltas} show the class conditional view of calibration changes vs. accuracy changes for all datasets and language models.
Note that for \tzeropp, prompts labeling the negative class have little-to-no change in accuracy after calibration.

\begin{table}[!ht]
\centering
\begin{tabular}{ l cc cc cc }
\toprule
& \multicolumn{2}{c}{Youtube (Accuracy)} & \multicolumn{2}{c}{SMS (F1)} & \multicolumn{2}{c}{Spouse (F1)} \\ \cmidrule(lr){2-3}\cmidrule(lr){4-5}\cmidrule(lr){6-7}
& Zero Shot & Prompted WS & Zero Shot &  Prompted WS & Zero Shot & Prompted WS \\ \midrule
WRENCH Benchmark    & -          & 94.9 (0.5) & - &  92.4 (0.5)         &    -        &  37.9 (2.8)          \\ \midrule
\tzeropp                & 54.1 (0.6) & \textbf{95.4 (0.4)} & 84.1 (1.7) & 91.4 (1.6)  & \textbf{60.6 (0.8)} & 44.9 (1.3) \\
InstructGPT Curie      & 52.8 (0.0) & 77.7 (1.1) & 0.0 (0.0)  & 0.0 (0.0)   & 0.0 (0.0)  & 0.0 (0.0)  \\
InstructGPT Babbage & 52.8 (0.0) & 69.2 (3.0) & 0.0 (0.0)  & 40.5 (10.3) & 33.8 (7.2) & 0.0 (0.0)  \\
InstructGPT Ada     & 52.8 (0.0) & 67.3 (1.2) & 0.0 (0.0)  & \textbf{94.7 (0.5)}  & 26.1 (1.2) & 0.0 (0.0)  \\
\bottomrule
\end{tabular}
\caption{The same performance metrics presented in Table \ref{tab:prompt_results_main} but with uncalibrated prompts.}
\label{tab:prompt_results_uncalibrated}
\vspace{-4mm}
\end{table}

%
%
\begin{table}[!ht]
\centering
\begin{tabular}{ll | crrrclc}
\toprule
Dataset &      Language Model &        CC &   Precision &      Recall &          F1 &         $\pm$F1 &       Acc. &      $\pm$Acc. \\
\midrule
\multirow{2}{*}{YouTube} &  \multirow{2}{*}{\tzeropp} & \checkmark &  61.5 (7.3) &  56.5 (2.6) &  48.0 (4.9) &   \textbf{+10.5 (5.0)} & 58.7 (2.4) &   \textbf{+4.7 (2.4)} \\
 &               &           & 50.6 (10.9) &  51.3 (0.6) &  37.5 (1.3) &   \textendash & 54.1 (0.6) &  \textendash \\ \midrule
\multirow{2}{*}{SMS} &   \multirow{2}{*}{\tzeropp} & \checkmark &  77.5 (3.9) &  90.3 (1.1) &  83.2 (2.4) &   -0.9 (2.1) & 95.0 (0.8) &  -0.5 (0.7) \\
&              &           &  82.4 (4.5) &  88.1 (4.3) &  84.1 (1.7) &   \textendash & 95.5 (0.5) &  \textendash \\ \midrule
\multirow{2}{*}{Spouse} &  \multirow{2}{*}{\tzeropp} & \checkmark & 37.3 (11.8) & 46.7 (14.8) & 41.5 (13.1) & -19.1 (13.6) & 92.7 (0.3) &   \textbf{+0.1 (0.7)} \\
&                 &           &  54.3 (2.2) &  69.7 (3.1) &  60.6 (0.8) &   \textendash & 92.6 (0.5) &  \textendash \\ \midrule
\multirow{2}{*}{YouTube} &   \multirow{2}{*}{InstructGPT Curie} & \checkmark &  26.4 (0.0) &  50.0 (0.0) &  34.6 (0.0) &    0.0 (0.0) & 52.8 (0.0) &   0.0 (0.0) \\
 &    &           &  26.4 (0.0) &  50.0 (0.0) &  34.6 (0.0) &   \textendash & 52.8 (0.0) &  \textendash \\ \midrule
\multirow{2}{*}{SMS} &   \multirow{2}{*}{InstructGPT Curie} & \checkmark &   0.0 (0.0) &   0.0 (0.0) &   0.0 (0.0) &    0.0 (0.0) & 86.6 (0.0) &   0.0 (0.0) \\
     &    &           &   0.0 (0.0) &   0.0 (0.0) &   0.0 (0.0) &   \textendash & 86.6 (0.0) &  \textendash \\ \midrule
\multirow{2}{*}{Spouse} &   \multirow{2}{*}{InstructGPT Curie} & \checkmark &  37.9 (1.8) &  74.5 (4.8) &  49.6 (1.0) &   \textbf{+49.6 (1.0)} & 87.7 (1.0) &  -4.2 (1.0) \\
  &  &           &   0.0 (0.0) &   0.0 (0.0) &   0.0 (0.0) &   \textendash & 91.9 (0.0) &  \textendash \\ \midrule
\multirow{2}{*}{YouTube} & \multirow{2}{*}{InstructGPT Babbage} & \checkmark &  81.4 (2.2) &  77.6 (3.2) &  77.2 (3.7) &   \textbf{+42.6 (3.7)} & 78.5 (3.0) &  \textbf{+25.7 (3.0)} \\
 &  &           &  26.4 (0.0) &  50.0 (0.0) &  34.6 (0.0) &   \textendash & 52.8 (0.0) &  \textendash \\ \midrule
\multirow{2}{*}{SMS} & \multirow{2}{*}{InstructGPT Babbage} & \checkmark &  20.8 (2.7) &  79.4 (5.4) &  32.2 (3.0) &  \textbf{ +32.2 (3.0)} & 52.0 (8.0) & -34.6 (8.0) \\
     &  &           &   0.0 (0.0) &   0.0 (0.0) &   0.0 (0.0) &   \textendash & 86.6 (0.0) &  \textendash \\ \midrule
\multirow{2}{*}{Spouse} & \multirow{2}{*}{InstructGPT Babbage} & \checkmark &  28.5 (1.1) &  75.3 (6.0) &  40.9 (0.9) &    \textbf{+7.1 (7.1)} & 82.5 (1.3) &  -6.1 (0.6) \\
  &  &           &  28.1 (5.8) & 43.7 (10.0) &  33.8 (7.2) &   \textendash & 88.5 (0.8) &  \textendash \\ \midrule
\multirow{2}{*}{YouTube} &    \multirow{2}{*}{InstructGPT Ada}& \checkmark &  40.6 (7.7) &  53.4 (1.9) &  43.4 (5.3) &    \textbf{+8.9 (5.3)} & 51.7 (2.4) &  -1.1 (2.4) \\
 &    &           &  26.4 (0.0) &  50.0 (0.0) &  34.6 (0.0) &   \textendash & 52.8 (0.0) &  \textendash \\ \midrule
\multirow{2}{*}{SMS} &     \multirow{2}{*}{InstructGPT Ada} & \checkmark &  15.3 (1.9) &  99.8 (0.2) &  26.3 (2.6) &   \textbf{+26.3 (2.6)} & 21.1 (7.7) & -65.5 (7.7) \\
     &    &           &   0.0 (0.0) &   0.0 (0.0) &   0.0 (0.0) &   \textendash & 86.6 (0.0) &  \textendash \\ \midrule
\multirow{2}{*}{Spouse} &     \multirow{2}{*}{InstructGPT Ada} & \checkmark &  10.6 (0.5) &  97.9 (0.6) &  19.1 (0.8) &   -7.0 (1.7) & 32.2 (3.9) & -23.2 (5.8) \\
  &     &           &  15.1 (0.8) &  96.1 (0.9) &  26.1 (1.2) &   \textendash & 55.4 (2.9) &  \textendash \\ 
\bottomrule
\end{tabular}
\caption{Performance impact of contextual calibration (CC) on all Zero Shot baseline models. Scores are the mean/standard error of 6 training replicates. Overall improvements due to calibration are in bold.}
\label{tab:zs_cc_all}
\end{table}

%
%
\begin{table}[!ht]
\centering
\begin{tabular}{lll rrrclc}
\toprule
Dataset &      Language Model &        CC &   Precision &      Recall &          F1 &         $\pm$F1 &       Acc. &      $\pm$Acc. \\
\midrule
YouTube &                \tzeropp & \checkmark &  92.6 (0.5) &  91.7 (0.5) &  91.9 (0.5) &   -3.5 (0.6) & 92.0 (0.5) &  -3.4 (0.6) \\
YouTube &                \tzeropp &           &  95.7 (0.4) &  95.2 (0.5) &  95.4 (0.4) &   \textendash & 95.4 (0.4) &  \textendash \\ \midrule
    SMS &                \tzeropp & \checkmark &  95.9 (2.5) &  88.1 (1.1) &  91.8 (1.6) &   \textbf{+0.3 (2.5)} & 97.9 (0.4) &  +0.2 (0.7) \\
    SMS &                \tzeropp &           &  91.6 (3.2) &  91.5 (0.8) &  91.4 (1.6) &   \textendash & 97.7 (0.5) &  \textendash \\ \midrule
 Spouse &                \tzeropp & \checkmark &  54.2 (1.8) &  75.4 (1.2) &  62.9 (0.8) &  \textbf{+18.0 (1.7)} & 92.8 (0.3) & \textbf{+10.0 (1.4)} \\
 Spouse &                \tzeropp &           &  30.7 (1.6) &  86.0 (2.8) &  44.9 (1.3) &   \textendash & 82.8 (1.2) &  \textendash \\ \midrule
YouTube &   InstructGPT Curie & \checkmark &  80.1 (1.0) &  77.1 (2.1) &  76.7 (2.3) &   \textbf{+0.8 (2.0)} & 77.7 (1.9) &  -0.1 (1.6) \\
YouTube &   InstructGPT Curie &           &  84.8 (0.6) &  76.4 (1.2) &  75.9 (1.3) &   \textendash & 77.7 (1.1) &  \textendash \\ \midrule
    SMS &   InstructGPT Curie & \checkmark & 60.6 (11.5) &  83.8 (4.4) &  65.7 (5.8) &  \textbf{+65.7 (5.8)} & 86.2 (3.9) &  -0.4 (3.9) \\
    SMS &   InstructGPT Curie &           &   0.0 (0.0) &   0.0 (0.0) &   0.0 (0.0) &   \textendash & 86.6 (0.0) &  \textendash \\ \midrule
 Spouse &   InstructGPT Curie & \checkmark &  29.5 (0.9) &  67.5 (2.0) &  41.0 (0.9) &  \textbf{+41.0 (0.9)} & 84.3 (0.7) &  -7.7 (0.7) \\
 Spouse &   InstructGPT Curie &           &   0.0 (0.0) &   0.0 (0.0) &   0.0 (0.0) &   \textendash & 91.9 (0.0) &  \textendash \\ \midrule
YouTube & InstructGPT Babbage & \checkmark &  85.8 (1.2) &  84.8 (1.4) &  84.9 (1.4) & \textbf{ +18.2 (4.3)} & 85.1 (1.3) & \textbf{+15.9 (3.6)} \\
YouTube & InstructGPT Babbage &           &  74.2 (3.4) &  68.2 (3.0) &  66.7 (3.5) &   \textendash & 69.2 (3.0) &  \textendash \\ \midrule
    SMS & InstructGPT Babbage & \checkmark &  13.4 (0.0) & 100.0 (0.0) &  23.6 (0.0) & -16.9 (10.3) & 13.4 (0.0) & -72.1 (2.4) \\
    SMS & InstructGPT Babbage &           & 48.9 (12.2) & 48.5 (15.1) & 40.5 (10.3) &   \textendash & 85.5 (2.4) &  \textendash \\ \midrule
 Spouse & InstructGPT Babbage & \checkmark &  24.2 (2.2) &  67.7 (5.6) &  34.9 (1.7) &  \textbf{+34.9 (1.7)} & 79.3 (1.9) & -12.6 (1.9) \\
 Spouse & InstructGPT Babbage &           &   0.0 (0.0) &   0.0 (0.0) &   0.0 (0.0) &   \textendash & 91.9 (0.0) &  \textendash \\ \midrule
YouTube &     InstructGPT Ada & \checkmark &  34.8 (8.4) &  50.1 (0.1) &  34.7 (0.2) &  -27.6 (1.9) & 52.9 (0.1) & -14.4 (1.2) \\ 
YouTube &     InstructGPT Ada &           &  77.5 (1.4) &  65.5 (1.3) &  62.3 (1.9) &   \textendash & 67.3 (1.2) &  \textendash \\ \midrule
    SMS &     InstructGPT Ada & \checkmark &  16.6 (1.3) &  99.5 (0.5) &  28.3 (1.8) &  -66.4 (1.9) & 30.4 (6.2) & -68.2 (6.2) \\
    SMS &     InstructGPT Ada &           &  98.7 (0.7) &  91.0 (0.8) &  94.7 (0.5) &   \textendash & 98.6 (0.1) &  \textendash \\ \midrule
 Spouse &     InstructGPT Ada & \checkmark &  16.8 (5.6) &  20.6 (8.1) &  17.7 (6.2) & \textbf{ +17.7 (6.2)} & 88.7 (1.4) &  -3.3 (1.4) \\
 Spouse &     InstructGPT Ada &           &   0.0 (0.0) &   0.0 (0.0) &   0.0 (0.0) &   \textendash & 91.9 (0.0) &  \textendash \\ 
\bottomrule
\end{tabular}
\caption{Performance impact of contextual calibration (CC) on all Prompted Weak Supervision models. Scores are the mean/standard error of 6 training replicates. Overall improvements due to calibration are in bold.}
\label{tab:pws_cc_all}
\end{table}

\subsection{WRENCH Labeling Function Prompts}

The complete set of translated WRENCH labeling functions are show in Tables \ref{tab:yt_prompt_templates}, \ref{tab:sms_prompt_templates}, and \ref{tab:spouse_prompt_templates}.

%
%

\label{section:wrench_lf_prompts}
\begin{table}[!ht]
\centering
\begin{tabular}{l|p{12cm}|c}
\toprule
Model & Prompt Template & Label \\
\midrule
\multirow{10}{*}{\tzeropp}
& \texttt{Does the following comment reference the speaker's channel or video?\textbackslash n\textbackslash n[TEXT]} & \texttt{SPAM} \\ 
& \texttt{Does the following comment ask you to subscribe to a channel?\textbackslash n\textbackslash n[TEXT]} & \texttt{SPAM} \\ 
& \texttt{Does the following comment have a URL?\textbackslash n\textbackslash n[TEXT]} & \texttt{SPAM} \\ 
& \texttt{Does the following comment ask the reader to do something?\textbackslash n\textbackslash n[TEXT]} & \texttt{SPAM} \\ 
& \texttt{Does the following comment talk about a song?\textbackslash n\textbackslash n[TEXT]} & \texttt{HAM} \\ 
& \texttt{Does the following comment contain the words "check out"? \textbackslash n\textbackslash n[TEXT]} & \texttt{SPAM} \\ 
& \texttt{Is the following comment fewer than 5 words?\textbackslash n\textbackslash n[TEXT]} & \texttt{HAM} \\ 
& \texttt{Does the following comment mention a person's name?\textbackslash n\textbackslash n[TEXT]} & \texttt{HAM} \\ 
& \texttt{Does the following comment express a very strong sentiment?\textbackslash n\textbackslash n[TEXT]} & \texttt{HAM} \\ 
& \texttt{Does the following comment express a subjective opinion?\textbackslash n\textbackslash n[TEXT]} & \texttt{HAM} \\ \midrule
\multirow{10}{*}{GPT-3} & \texttt{Q: Does the following comment "[TEXT]" reference the speaker's channel or video?\textbackslash nA:} & \texttt{SPAM} \\
& \texttt{Q: Does the following comment "[TEXT]" ask you to subscribe to a channel?\textbackslash nA:} & \texttt{SPAM} \\
& \texttt{Q: Does the following comment "[TEXT]" have a URL?\textbackslash nA:} & \texttt{SPAM} \\
& \texttt{Q: Does the following comment "[TEXT]" ask the reader to do something?\textbackslash nA:} & \texttt{SPAM} \\
& \texttt{Q: Does the following comment "[TEXT]" talk about a song?\textbackslash nA:} & \texttt{HAM} \\
& \texttt{Q: Does the following comment "[TEXT]" contain the words "check out"?\textbackslash nA:} & \texttt{SPAM} \\
& \texttt{Q: Is the following comment "[TEXT]" fewer than 5 words?\textbackslash nA:} & \texttt{HAM} \\
& \texttt{Q: Does the following comment "[TEXT]" mention a person's name?\textbackslash nA:} & \texttt{HAM} \\
& \texttt{Q: Does the following comment "[TEXT]" express a very strong sentiment?\textbackslash nA:} &\texttt{HAM}\\
& \texttt{Q: Does the following comment "[TEXT]" express a subjective opinion?\textbackslash nA:} & \texttt{HAM} \\
\bottomrule
\end{tabular}
\caption{YouTube labeling function prompts with class labels \texttt{HAM} = 0, \texttt{SPAM} = 1. A label map transforms text completions to class labels, where "yes" emits the value denoted in the label column and "no" emits ABSTAIN.}
\label{tab:yt_prompt_templates}
\end{table} 

%
%
\begin{table}[!ht]
\centering
\begin{tabular}{l|p{12cm}|c}
\toprule
Model & Prompt Template & Label \\ \midrule
\tzeropp & \texttt{Does the following text message contain the words "[KEYWORDS]"?\textbackslash n\textbackslash n[TEXT]} & \\ \midrule
GPT-3 & \texttt{Q: Does the following text message "[TEXT]" contain the words "[KEYWORDS]"?\textbackslash nA:} & \\ \midrule
\multirow{9}{*}{\texttt{[KEYWORDS]}} & \textit{??1.50, ??500, ??5000, call for offer, cash prize, chat date, chat to, childporn, credits, dating call, direct, expires now, fantasies call, free phones, free price, free ringtones, free sex, free tone, guaranteed free, guaranteed gift, hard live girl, important lucky, inviting friends, latest, latest offer, message call, new mobiles, no extra, password, please call, sms reply, unlimited calls, urgent award guaranteed, urgent prize, voucher claim, welcome reply, win shopping, winner reward, won call, won cash, won cash prize, won claim }  & \texttt{SPAM} \\ \cmidrule{2-3}
 & \textit{I, I can did, I it, I miss, I used to, adventuring, amrita, can't talk, did u got, do you, fb, goodo, hee hee, i'll, jus, link, maggi, mine, my kids, noisy, praying, shit, should I, thanks, that's fine, thats nice, u how 2, we will, where are, wtf, your I} & \texttt{HAM} \\
\bottomrule
\end{tabular}
\caption{SMS Labeling function prompts with class labels \texttt{HAM} = 0, \texttt{SPAM} = 1 that are defined by individual \texttt{[KEYWORDS]}. A label map transforms text completions to class labels, where "yes" emits the value denoted in the label column and "no" emits ABSTAIN.}
\label{tab:sms_prompt_templates}
\end{table}

%
%
\begin{table}[!ht]
\centering
\begin{tabular}{l|p{12cm}|c}
\toprule
Model & Prompt Template & Label \\
\midrule
\multirow{11}{*}{\tzeropp} &  \texttt{Context: [TEXT]\textbackslash n\textbackslash nIs there any mention of "spouse" between the entities [PERSON1] and [PERSON2]?} &  \texttt{SPOUSE} \\
& \texttt{Context: [TEXT]\textbackslash n\textbackslash nIs there any mention of "spouse" before the entity [PERSON1]?} & \texttt{SPOUSE} \\
& \texttt{Context: [TEXT]\textbackslash n\textbackslash nIs there any mention of "spouse" before the entity [PERSON2]?} & \texttt{SPOUSE} \\
& \texttt{Context: [TEXT]\textbackslash n\textbackslash nDo [PERSON1] and [PERSON2] have the same last name?} & \texttt{SPOUSE} \\
& \texttt{Context: [TEXT]\textbackslash n\textbackslash nDid [PERSON1] and [PERSON2] get married?} & \texttt{SPOUSE} \\
& \texttt{Context: [TEXT]\textbackslash n\textbackslash nAre [PERSON1] and [PERSON2] family members?} & \texttt{NOT\_SPOUSE} \\
& \texttt{Context: [TEXT]\textbackslash n\textbackslash nIs [PERSON1] said to be a family member?} & \texttt{NOT\_SPOUSE} \\
& \texttt{Context: [TEXT]\textbackslash n\textbackslash nIs [PERSON2] said to be a family member?} & \texttt{NOT\_SPOUSE} \\
& \texttt{Context: [TEXT]\textbackslash n\textbackslash nAre [PERSON1] and [PERSON2] dating?} & \texttt{NOT\_SPOUSE} \\
& \texttt{Context: [TEXT]\textbackslash n\textbackslash nAre [PERSON1] and [PERSON2] co-workers?}  & \texttt{NOT\_SPOUSE} \\
& \texttt{Are [PERSON1] and [PERSON2] married?} & \texttt{SPOUSE} \\
\midrule
\multirow{11}{*}{GPT-3} & \texttt{Context: "[TEXT]"\textbackslash nQ: Is there any mention of "spouse" between the entities [PERSON1] and [PERSON2]?\textbackslash nA}: & \texttt{SPOUSE}\\
& \texttt{Context: "[TEXT]"\textbackslash nQ: Is there any mention of "spouse" before the entity [PERSON1]?\textbackslash nA}: & \texttt{SPOUSE} \\
& \texttt{Context: "[TEXT]"\textbackslash nQ: Is there any mention of "spouse" before the entity [PERSON2]?\textbackslash nA}: & \texttt{SPOUSE} \\
& \texttt{Context: "[TEXT]"\textbackslash nQ: Do [PERSON1] and [PERSON2] have the same last name?\textbackslash nA}: & \texttt{SPOUSE} \\
& \texttt{Context: "[TEXT]"\textbackslash nQ: Did [PERSON1] and [PERSON2] get married?\textbackslash nA}: & \texttt{SPOUSE} \\
& \texttt{Context: "[TEXT]"\textbackslash nQ: Are [PERSON1] and [PERSON2] family members?\textbackslash nA}: & \texttt{NOT\_SPOUSE} \\
& \texttt{Context: "[TEXT]"\textbackslash nQ: Is [PERSON1] said to be a family member?\textbackslash nA}: & \texttt{NOT\_SPOUSE} \\
& \texttt{Context: "[TEXT]"\textbackslash nQ: Is [PERSON2] said to be a family member?\textbackslash nA}: & \texttt{NOT\_SPOUSE} \\
& \texttt{Context: "[TEXT]"\textbackslash nQ: Are [PERSON1] and [PERSON2] dating?\textbackslash nA}: & \texttt{NOT\_SPOUSE} \\
& \texttt{Context: "[TEXT]"\textbackslash nQ: Are [PERSON1] and [PERSON2] co-workers?\textbackslash nA}: & \texttt{NOT\_SPOUSE} \\ 
& \texttt{Q: Are [PERSON1] and [PERSON2] married?\textbackslash nA}: & \texttt{SPOUSE} \\ 
\bottomrule
\end{tabular}
\caption{Spouse labeling function prompts with class labels \texttt{NOT\_SPOUSE} = 0, \texttt{SPOUSE} = 1. A label map transforms text completions to class labels, where "yes" emits the value denoted in the label column and "no" emits ABSTAIN.}
\label{tab:spouse_prompt_templates}
\end{table}

\subsection{Labeling Function Diversity}
\label{sec:diversity}

\begin{figure}[!ht]
    \centering
    \includegraphics[width=1.0\textwidth]{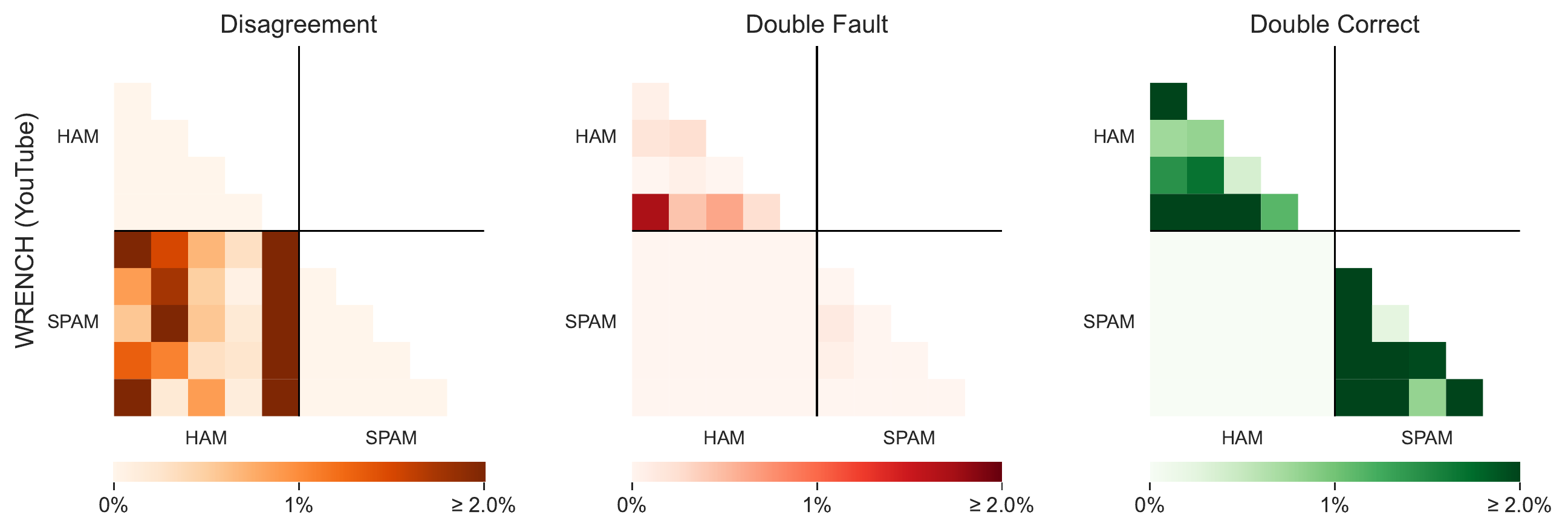}
    \includegraphics[width=1.0\textwidth]{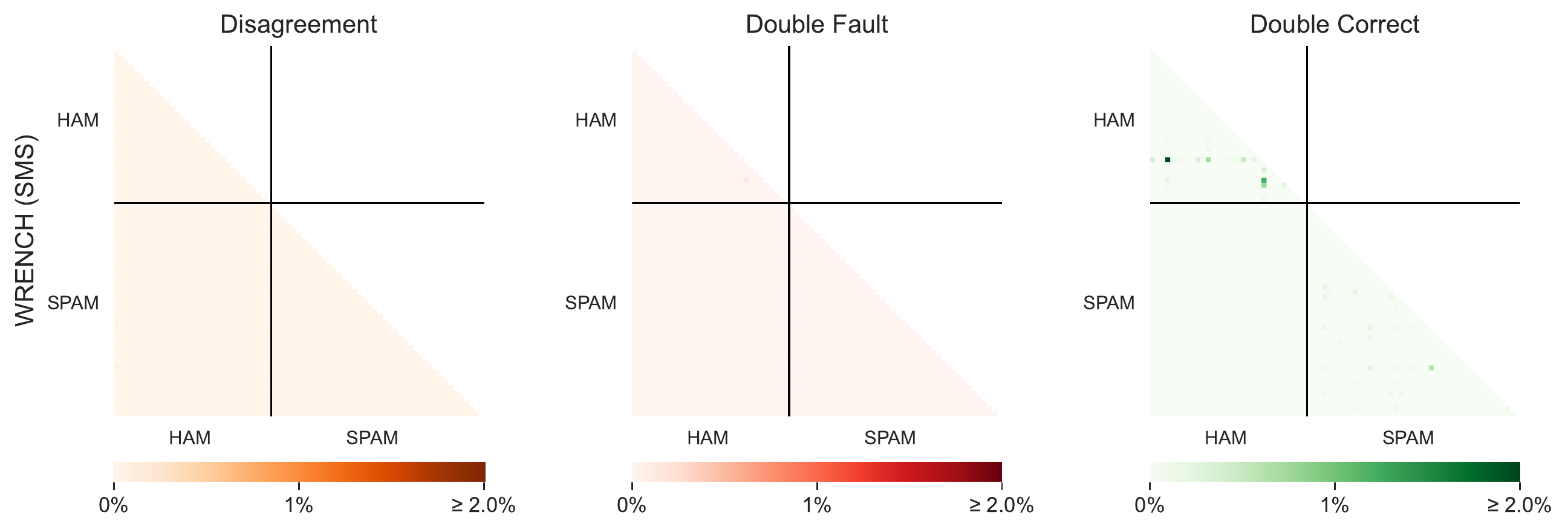}
    \includegraphics[width=1.0\textwidth]{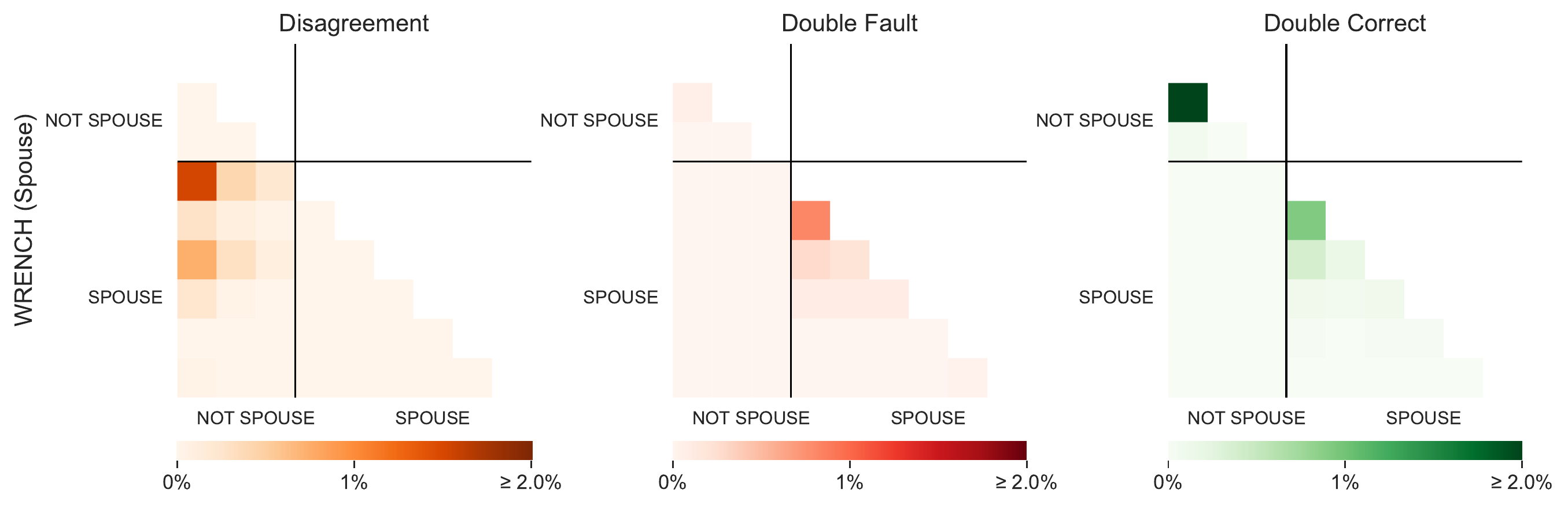}
    \caption{Diversity measures for the WRENCH Benchmark labeling function set. Here rules have very low coverage (i.e., rules typically vote on less that 2\% of the training set) but have high precision. SMS and Spouse have very low overall disagreement levels. YouTube has higher disagreement, but only limited cases where both labeling functions make correlated errors (double fault).  }
    \label{fig:wrench_lf_diversity}
\end{figure}

\begin{figure}[!ht]
    \centering
    \includegraphics[width=1.0\textwidth]{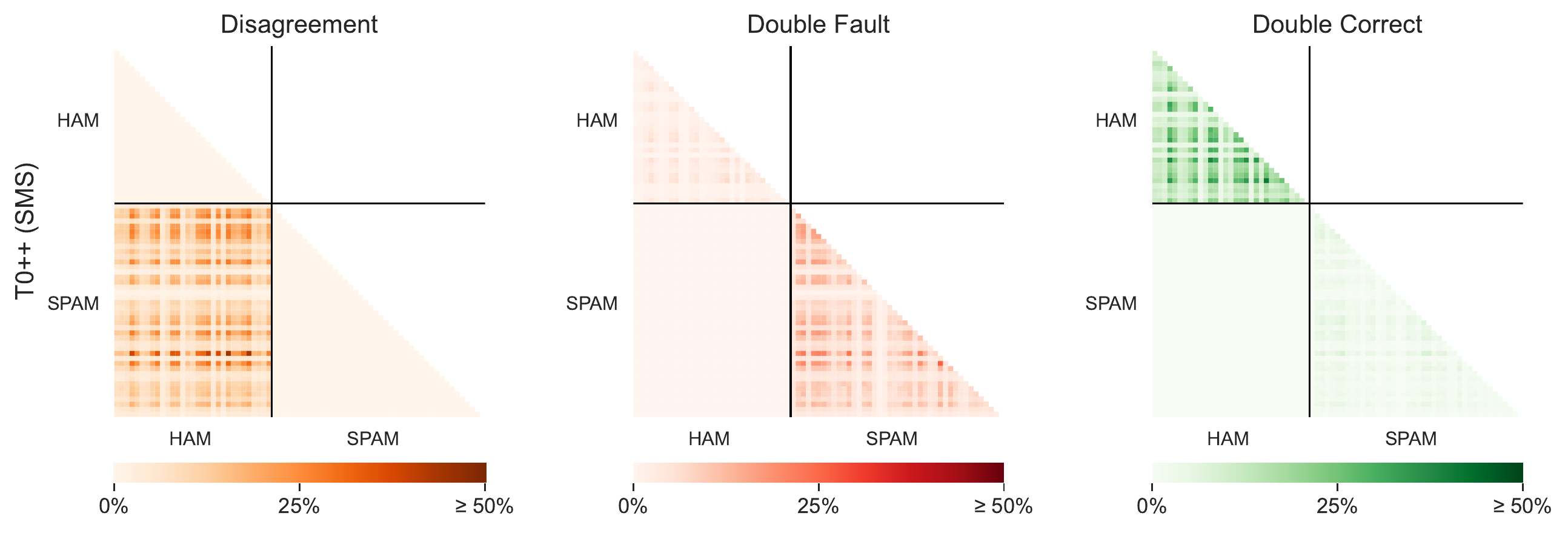}
    \includegraphics[width=1.0\textwidth]{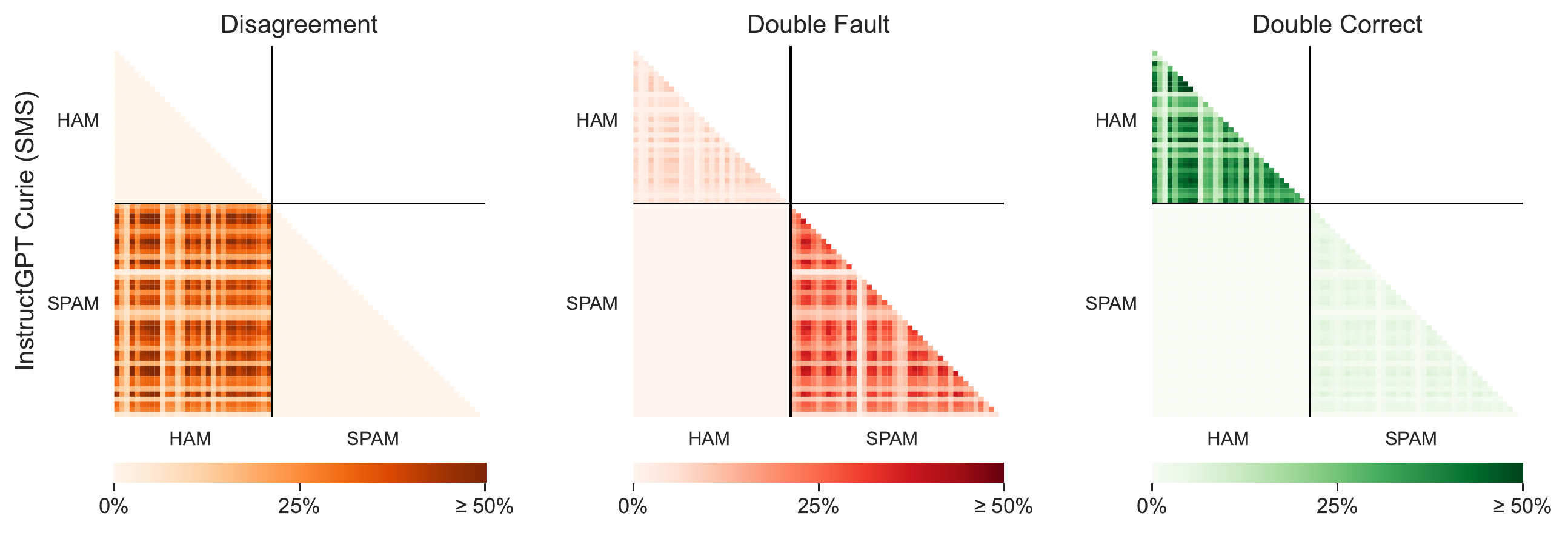}
    \includegraphics[width=1.0\textwidth]{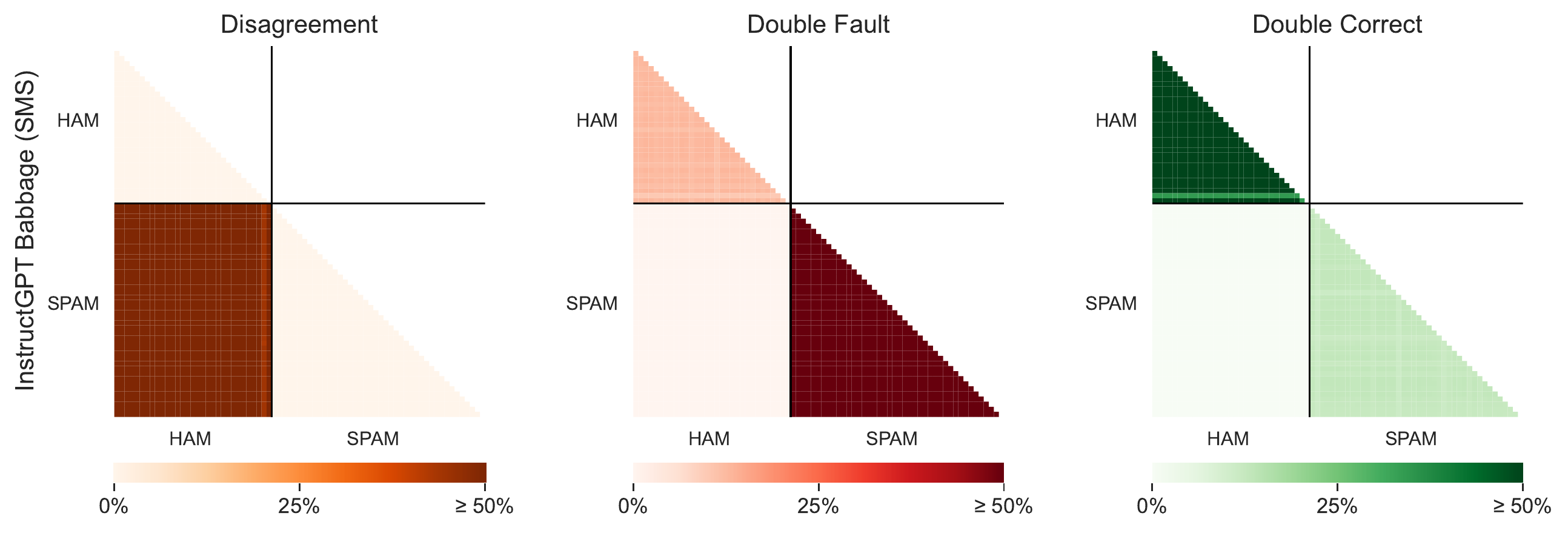}
    \includegraphics[width=1.0\textwidth]{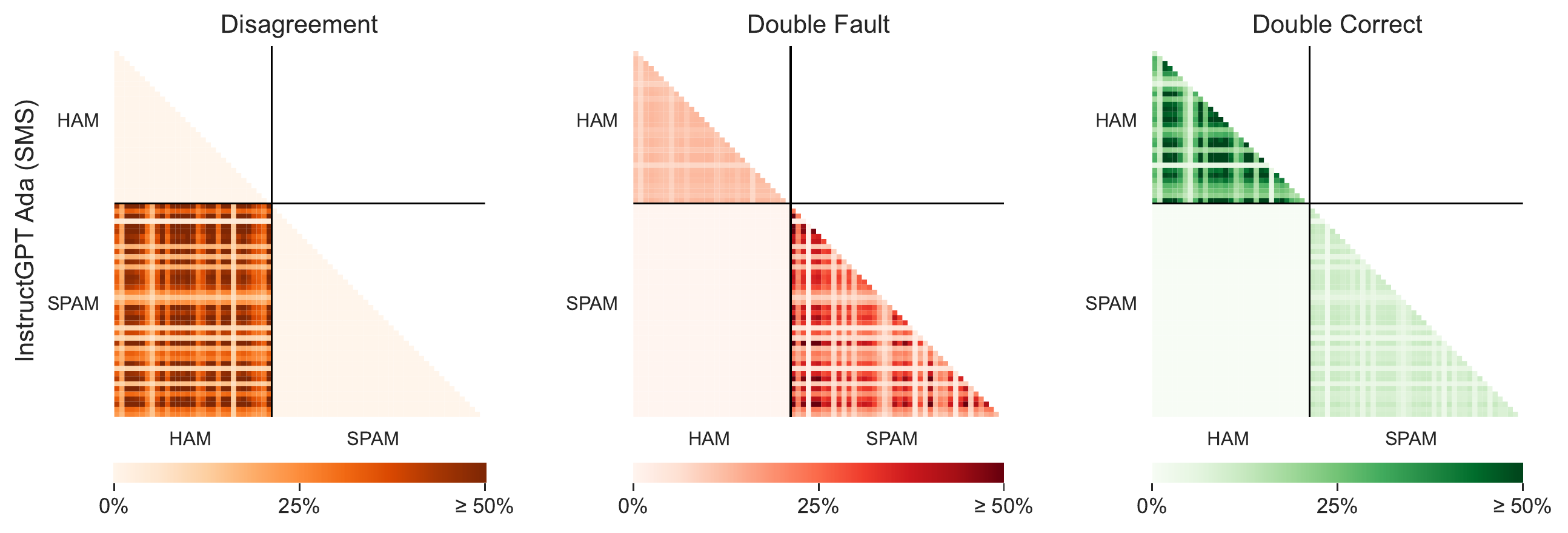}
    \caption{SMS prompted labeling function diversity measures. Color intensity represents the percentage of training examples labeled by a pair of prompts.}
    \label{fig:sms_lf_diversity}
\end{figure}

\begin{figure}[!ht]
    \centering
    \includegraphics[width=1.0\textwidth]{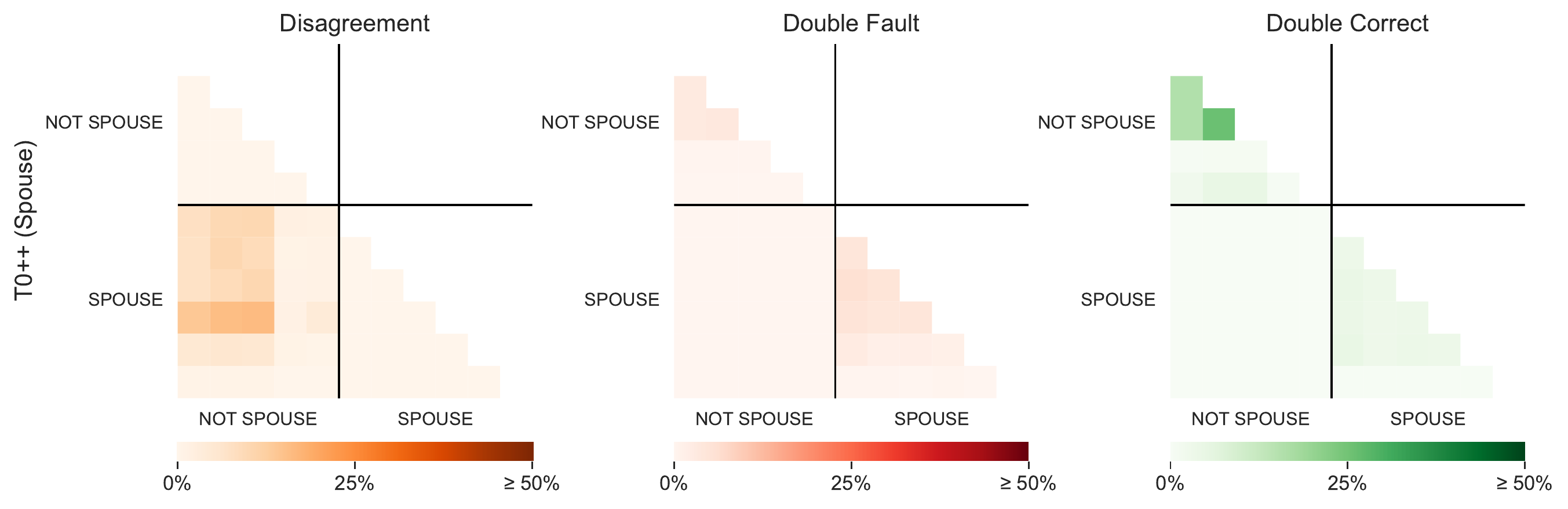}
    \includegraphics[width=1.0\textwidth]{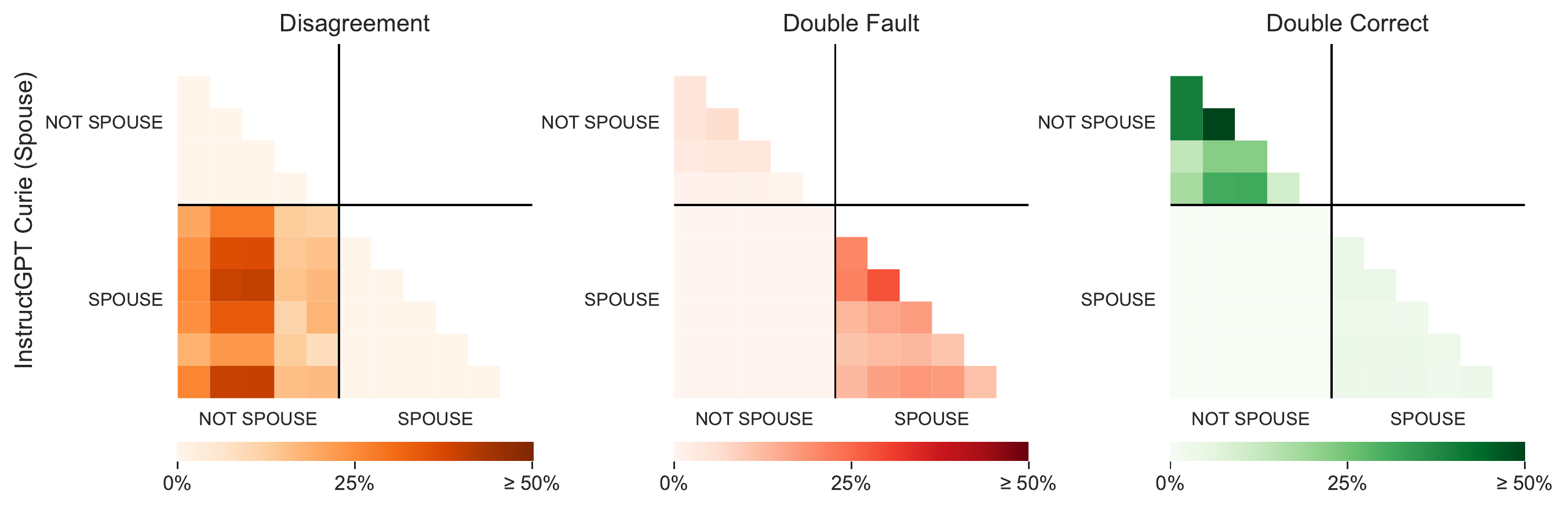}
    \includegraphics[width=1.0\textwidth]{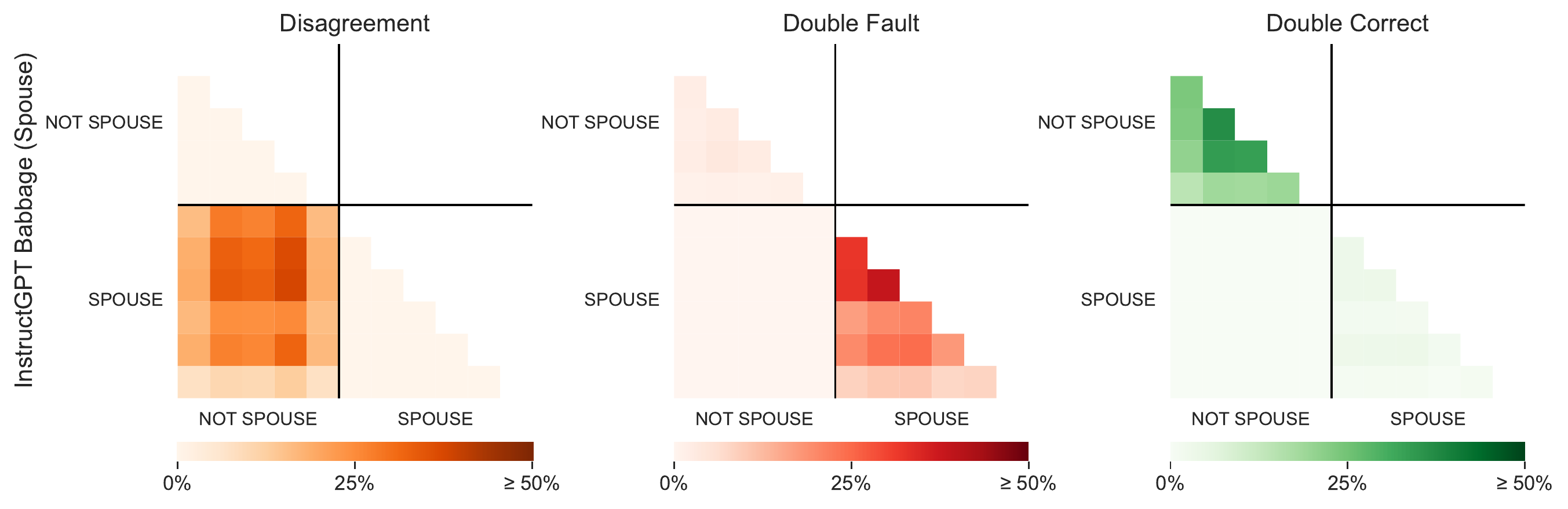}
    \includegraphics[width=1.0\textwidth]{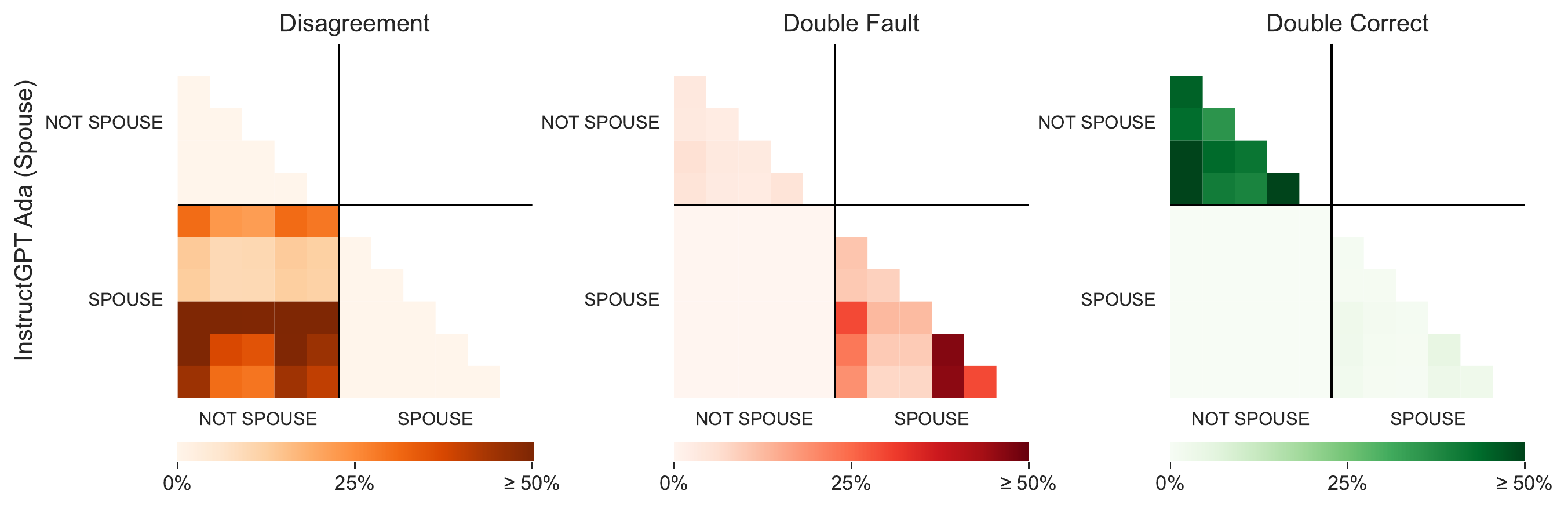}
    \caption{Spouse prompted labeling function diversity measures. Color intensity represents the percentage of training examples labeled by a pair of prompts.}
    \label{fig:spouse_lf_diversity}
\end{figure}

Figure \ref{fig:wrench_lf_diversity} shows a heatmap view of diversity metrics for the original WRENCH labeling functions. 
Figures \ref{fig:sms_lf_diversity} and \ref{fig:spouse_lf_diversity} show diversity measures for the SMS and Spouse datasets respectively.

\begin{figure}[!ht]
\centering
\includegraphics[width=1.0\textwidth]{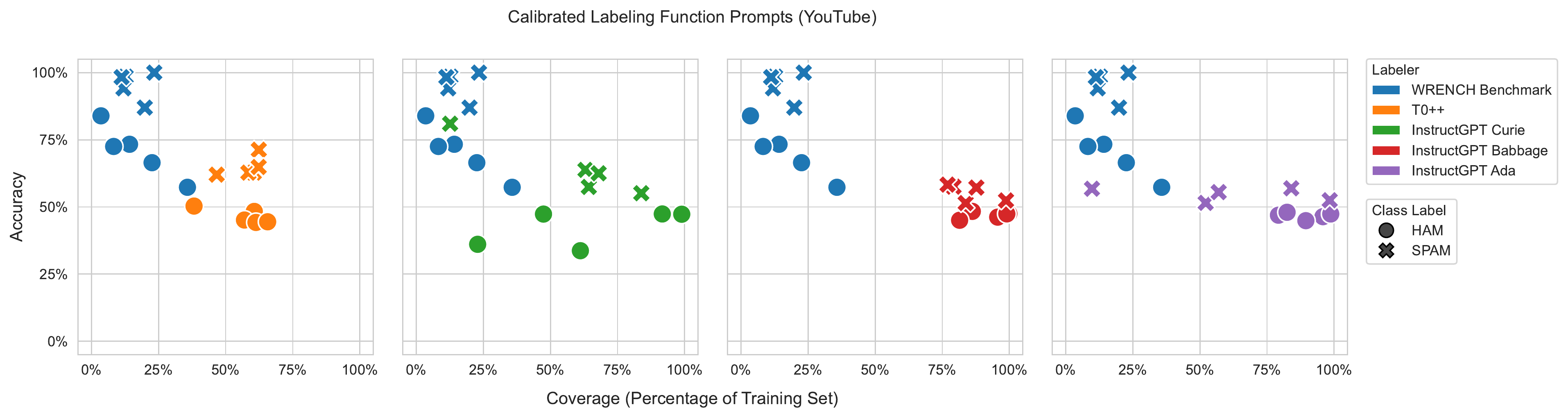}
\includegraphics[width=1.0\textwidth]{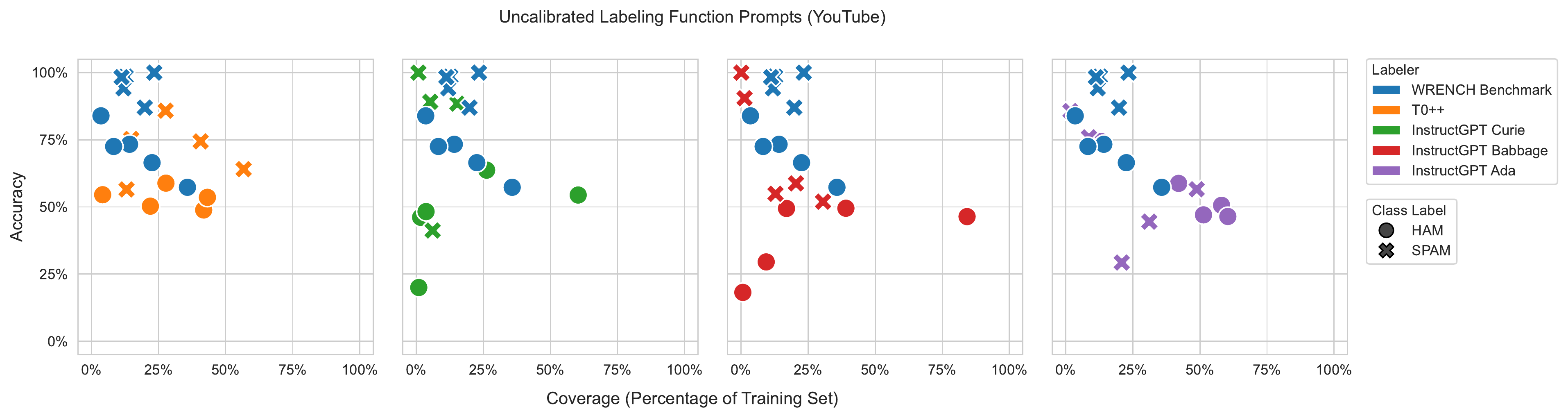}
\caption{YouTube prompted labeling function accuracy vs. coverage scatter plots. The top figure is calibrated using contextual calibration and the bottom is uncalibrated. Colors correspond to the language models used for labeling and marker style indicates class label. }
\label{fig:youtube_lf_acc_cov}
\end{figure}

\begin{figure}[!ht]
    \centering
    \includegraphics[width=1.0\textwidth]{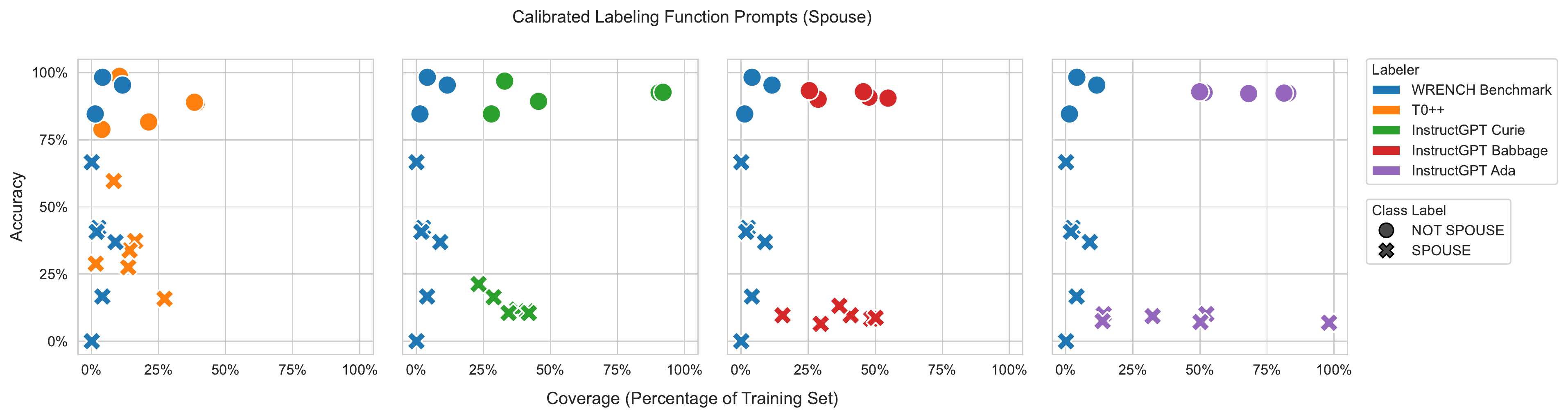}
    \includegraphics[width=1.0\textwidth]{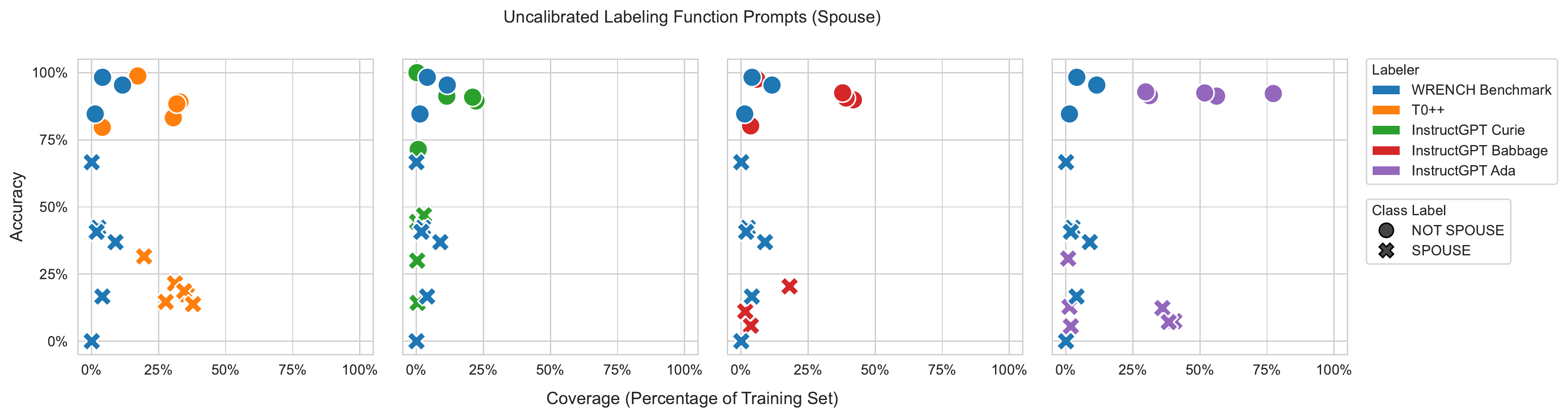}
    \caption{Spouse prompted labeling function accuracy vs. coverage scatter plots. The top figure is calibrated using contextual calibration and the bottom is uncalibrated. Colors correspond to the language models used for labeling and marker style indicates class label. }
    \label{fig:spouse_lf_acc_cov}
\end{figure}

\end{document}